\documentclass{article}

\usepackage{algorithm,algorithmic}
\usepackage{amsmath}
\usepackage{multirow}

\usepackage[preprint]{neurips_2025}

\usepackage[utf8]{inputenc} % allow utf-8 input
\usepackage[T1]{fontenc}    % use 8-bit T1 fonts
\usepackage{hyperref}       % hyperlinks
\usepackage{url}            % simple URL typesetting
\usepackage{booktabs}       % professional-quality tables
\usepackage{amsfonts}       % blackboard math symbols
\usepackage{nicefrac}       % compact symbols for 1/2, etc.
\usepackage{microtype}      % microtypography

\usepackage{subcaption}
\usepackage{graphicx}
\usepackage{pifont}
\usepackage{tcolorbox}
\usepackage{colortbl}
\usepackage{enumitem}
\usepackage{marvosym}
\definecolor{linkcolor}{rgb}{0.956,0.298,0.235} 
\hypersetup{
	colorlinks=true,
	linkcolor=linkcolor,
	filecolor=blue,      
	urlcolor=magenta,
	citecolor=black,
}
\newcommand{\xmark}{\ding{55}}
\definecolor{myred}{RGB}{220, 50, 47}
\definecolor{myblue}{RGB}{38, 139, 210}
\newcommand{\PeRLcolor}{\textcolor{myred}{P}\textcolor{myblue}{e}RL}

\newcommand{\PeRL}{PeRL}

\newcommand{\ours}{Permutation-Enhanced Reinforcement Learning for Interleaved Vision-Language Reasoning}
\newcommand{\OURS}{\PeRL}

\title{\raisebox{-0.27\height}{\includegraphics[width=0.35in]{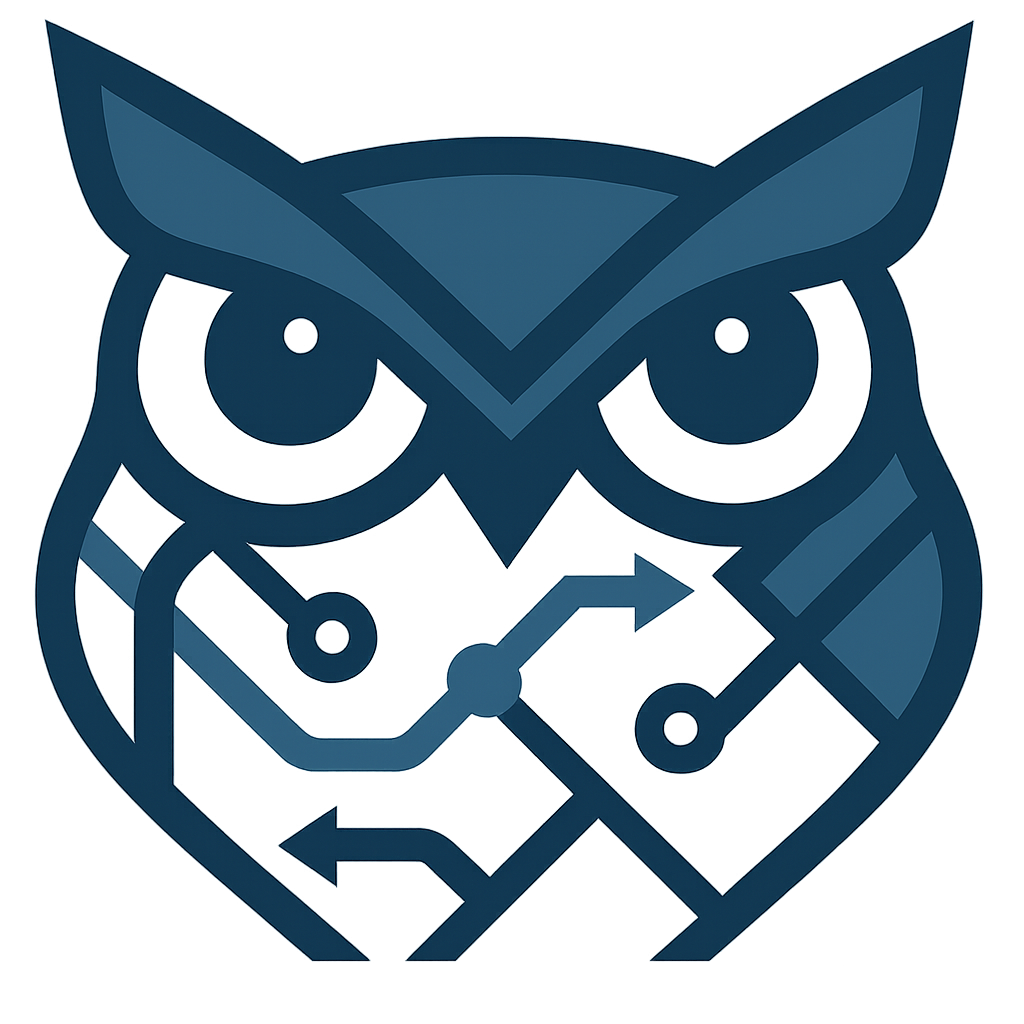}}~\PeRLcolor: \ours}
\author{
  Yizhen Zhang$^{\phi\pi}$\thanks{Equal contribution. Work done during the internship of Yizhen, Shuoshuo, Haoling, Zhong-Zhi and Jie at Microsoft Research.\Letter~:\texttt{zhangyizhen24@mails.tsinghua.edu.cn}}~~~~
  Yang Ding$^{\phi\,*}$~~~~
  Shuoshuo Zhang$^{\phi\pi\,*}$~~~~\\
  \textbf{Xinchen Zhang$^{\phi}$~~~~
  Haoling Li$^{\phi\pi}$~~~~
  Zhong-Zhi Li$^{\rho\pi}$~~~~
  Peijie Wang$^{\rho}$} \\
  \textbf{Jie Wu$^{\phi\pi}~~~~
  $Lei Ji$^{\pi}\thanks{Corresponding authors.}$~~~~
  Yelong Shen$^{\pi}$~~~~
  Yujiu Yang$^{\phi\dagger}$~~~~
  Yeyun Gong$^{\pi}$}  \\
$^\phi$Tsinghua University \quad
$^\pi$Microsoft \quad 
$^\rho$CASIA \\
\url{https://github.com/alchemistyzz/PeRL}
}

\begin{document}

\maketitle

\begin{abstract}

%wide real applications require general interleaved format to handle both single and multiple images. Face challenges of both spatial and positional reasoning. Existing multimodal RL for single-image reasoning only aiming spatial reasoning. explore and exploit. Diverse Rollout to explore. entropy vanish, or advantage vanish. 1) add gaussian/rotate noisy for single-image spatial diversity 2) permutation for multi-image positional diveristy. Sample effective Rollouts to exploit.

Inspired by the impressive reasoning capabilities demonstrated by reinforcement learning approaches like DeepSeek-R1, recent emerging research has begun exploring the use of reinforcement learning (RL) to enhance vision-language models (VLMs) for multimodal reasoning tasks. However, most existing multimodal reinforcement learning approaches remain limited to spatial reasoning within single-image contexts, yet still struggle to generalize to more complex and real-world scenarios involving multi-image positional reasoning, where understanding the relationships across images is crucial. To address this challenge, we propose a general reinforcement learning approach \OURS~tailored for interleaved multimodal tasks, and a multi-stage strategy designed to enhance the exploration-exploitation trade-off, thereby improving learning efficiency and task performance. 
Specifically, we introduce permutation of image sequences to simulate varied positional relationships to explore more spatial and positional  diversity. Furthermore, we design a rollout filtering mechanism for resampling to focus on trajectories that contribute most to learning optimal behaviors to exploit learned policies effectively. We evaluate our model on 5 widely-used multi-image benchmarks and 3 single-image benchmarks. Our experiments confirm that \OURS~trained model consistently surpasses R1-related and interleaved VLM baselines by a large margin, achieving state-of-the-art performance on multi-image benchmarks, while preserving comparable performance on single-image tasks.

% \vspace{1em}
% \begin{flushright}
% \textit{``Confusion is the beginning of wisdom.''} \\
% \textit{--- Socrates}
% \end{flushright}
% \vspace{1em}
\end{abstract}
\begin{figure}[ht]
% \vskip -1in
\begin{center}
\centerline{
\includegraphics[width=1.0\linewidth]{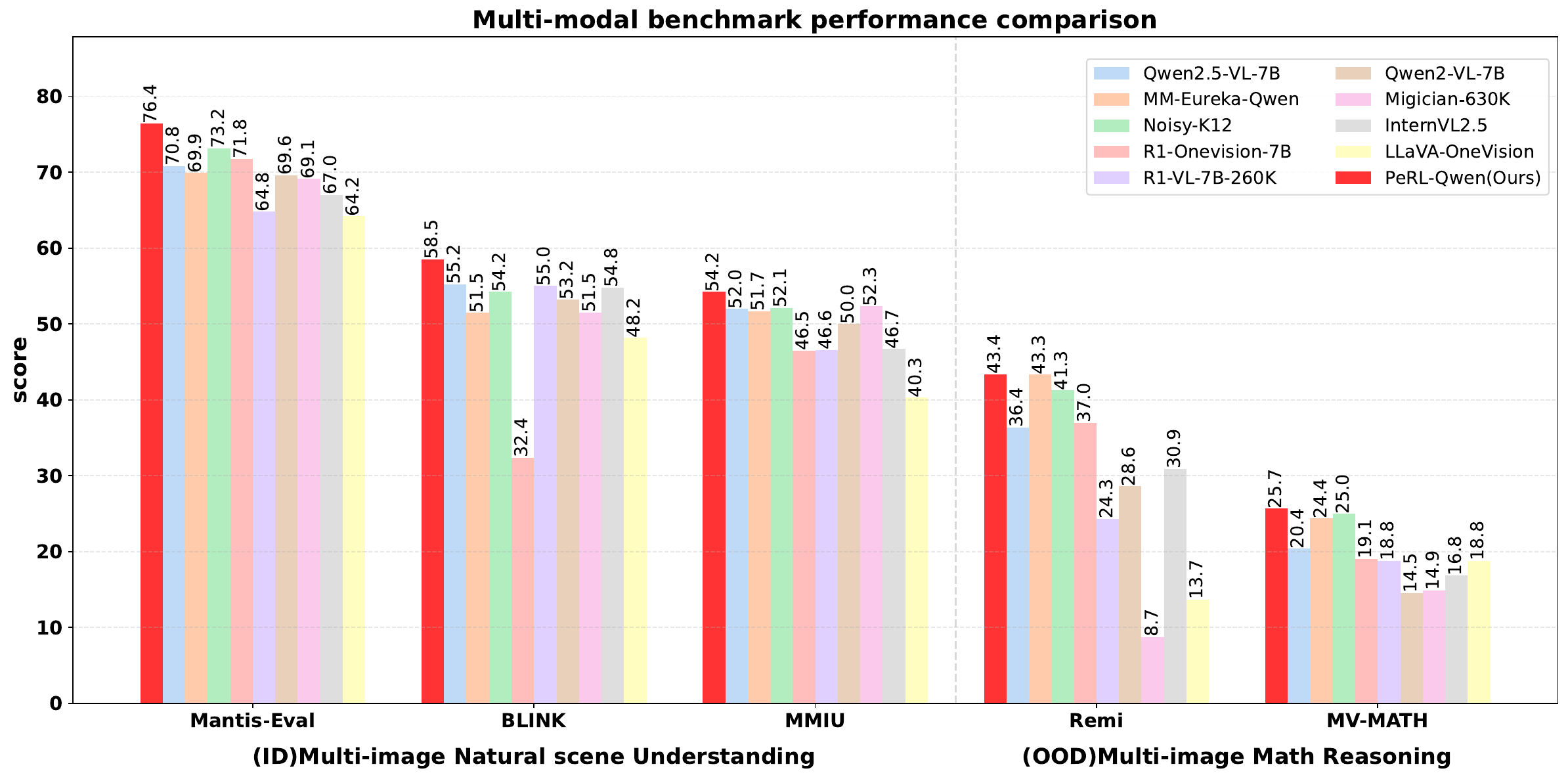}}
\caption{Compare \OURS~against baseline approaches, including instruction-tuned or R1-like reinforcement learning models, across both in-domain and out-of-domain multi-image benchmarks.
   }
\label{fig:multi_modal_benchmark}
\end{center}
\vskip -0.2in
\end{figure}

\section{Introduction}
%Real-world applications often demand multi-image understanding capability for comprehensive tasks, which human naturally has. Consequently, several multi-image benchmarks \cite{wang2024muirbench} have been released to contribute the community to study this comprehensive and applicable tasks. Specifically, multi-image multimodal benchmarks ranges from a broad range of tasks across various multimodal relationships including 1) low-level visual information features such as comparing illumination or saturation; 2) high-level relationships among objects such as emotional associations; 3) temporal orders; 4) 2D/3D spatial relationships such as rotation, symmetry or different camera positions. Furthermore, the interleaved format\cite{alayrac2022flamingo} of multiple images and text provide a unified template capable of covering both single-image and multi-image scenarios, where the input text and images  can be organized in arbitrary forms with flexible and dynamic interactions. 

Real-world applications often require the ability to understand and reason over multiple images—a capability that humans naturally possess. In response, several multi-image benchmarks (e.g., \cite{fu2024blink,jiang2024mantis,meng2024mmiu,wang2024muirbench}) have been introduced to support the study of such comprehensive and practically relevant tasks within the research community. Specifically, multi-image multimodal benchmarks span a wide range of tasks involving diverse multimodal relationships, including: 1)
Low-level visual features, such as comparing illumination, color saturation, or texture; 2)
High-level semantic relationships, such as emotional associations or conceptual similarities between objects; 3)
Temporal reasoning, involving the understanding of event sequences or changes over time; 4) 2D/3D spatial relationships, including rotation, symmetry, and variations in camera viewpoints.
Furthermore, the interleaved format of images and text, as introduced in works like \cite{alayrac2022flamingo}, provides a unified and flexible template that supports both single-image and multi-image scenarios. This format allows for arbitrary arrangements of textual and visual inputs, enabling dynamic and context-aware interactions that more closely resemble real-world reasoning tasks.

%However, despite the advanced performance with the progressive development of the VLMs in single-image tasks, the investigation of VLMs understanding multi-image tasks remains underexplored. Inspired by this, \cite{jiang2024mantis,li2024llava1} constructed multi-image instruction-tuning datasets to further post-train the general VLMs to tackle the multi-image tasks. Simultaneously, VLMs such as Flamingo\cite{alayrac2022flamingo}, LLava-interleave\cite{li2024llava1}, Mantis-Idefics\cite{}, have been trained on interleaved image-text data for multi-image understanding capability. Despite this, these interleaved VLMs still struggle in tackling the reasoning across images\cite{tian2025identifying}. Moreover, the use of reinforcement learning (RL) to enhance VLMs for multimodal reasoning tasks has emerged recently\cite{meng2025mm} and shown promising potential to incentive the multimodal reasoning for single-image tasks. Nonetheless, these models still suffer from lack of reasoning across images, especially for hard tasks that require a holistic understanding of multiple images.
Despite the impressive progress of vision-language models (VLMs) on single-image tasks, their ability to understand and reason over multiple images remains significantly underexplored. Motivated by this gap, recent works such as \cite{jiang2024mantis, li2024llava1,li2025migician} have constructed multi-image instruction-tuning datasets to further post-train general-purpose VLMs for multi-image tasks. In parallel, models like Flamingo \cite{alayrac2022flamingo}, LLaVA-Interleave \cite{li2024llava1}, Migician\cite{li2025migician} and Mantis-Idefics\cite{jiang2024mantis} have been trained on interleaved image-text data to enhance their multi-image understanding capabilities.
However, despite these efforts, interleaved VLMs still struggle with reasoning across images, as highlighted in \cite{tian2025identifying}. Their limitations become especially apparent in tasks that require positional alignment, temporal sequencing, or holistic cross-image understanding. Meanwhile, reinforcement learning (RL) has recently emerged as a promising direction for enhancing multimodal reasoning in VLMs \cite{meng2025mm}, showing encouraging results on single-image tasks. Nonetheless, these RL-augmented models also fall short when applied to more complex multi-image scenarios, particularly those demanding deep reasoning across spatial and semantic contexts.

%Concretely, these multi-image scenarios pose challenges in both spatial reasoning (understanding object relationships within an image) and positional reasoning (understanding relationships across multiple images). Specifically, as shown in Fig. \ref{fig:case}, after simply switching the position of the images, the answer of either the general model (Qwen-2.5-VL-7B) or instruction-tuned model (Migician) becomes incorrect. Qwen model failed to recognize the color of each image separately, which mentioned \emph{bright yellow and blue hues} in both images. Migician model is unable to associate the information to the right image, which explained \emph{red and green pattern} correctly but associate with the wrong image. 
Concretely, multi-image scenarios introduce challenges in both spatial reasoning—understanding object relationships within a single image—and positional reasoning—understanding relationships across multiple images. As illustrated in Fig. \ref{fig:case}, simply altering the order of input images can lead to incorrect predictions from both general-purpose and instruction-tuned vision-language models.
For instance, the general model Qwen-2.5-VL-7B\cite{bai2025qwen2} fails to distinguish the visual attributes of each image independently, incorrectly referencing “bright yellow and blue hues” for both images. Meanwhile, the instruction-tuned model Migician\cite{li2025migician} correctly identifies the “red and green pattern” but misattributes it to the wrong image, indicating a failure in positional alignment. These examples highlight the limitations of current VLMs in maintaining consistent and grounded reasoning across interleaved multi-image inputs.

%To deal with this, we investigate interleaved multimodal reasoning through reinforcement learning and designed a simple yet effective strategy to overcome the positional bias. In details, we first develop a multi-stage data processing pipeline to variant the data sample through image reordering. Since the semantic may change after the permutation, we alter the text accordingly to rephrase the sample to keep consistency. However, we observed an extreme imbalanced difficulty of these training samples. To efficiently reinforcement learning, we further introduce a rollout filtering mechanism cooperated with diversified advantage estimation to stabilize and effectively train our \OURS~model.

% from the data argumentation which takes each sample independently, we group each case with all permutated samples together to further calculate the advantage with diversified rollouts. This makes it more robust to position change. Besides, //////////////

To address these challenges, we explore interleaved multimodal reasoning through reinforcement learning and design a simple yet effective strategy to mitigate positional bias. Specifically, we develop a multi-stage data processing pipeline that generates diverse training samples by permuting the order of input images. Since such permutations may alter the semantic alignment between text and images, we correspondingly rephrase the textual input to maintain semantic consistency.
However, we observe a significant imbalance in the difficulty levels of the resulting training samples leading to inefficient reinforcement learning. To mitigate this, we introduce a rollout filtering mechanism which helps stabilize training and improve learning efficiency. This approach allows our proposed model, Qwen-\OURS, to focus on informative trajectories and better generalize across complex multi-image reasoning tasks.

%We have summarized the following contributions:
%\begin{itemize}
%  \item We introduce an interleaved multimodal model MMRL trained through reinforcement learning, which achieves strong multimodal reasoning capability for both single-image and multi-image tasks.
%  \item We propose a simple yet effective strategy to tackle the multi-image coreference problem for an effective training, and resample training data through difficulty calculation for an efficient training. 
%\end{itemize} 

We summarize our main contributions as follows:

\begin{itemize} 
% \item We introduce \OURS, an interleaved multimodal reinforcement learning framework that achieves strong reasoning capabilities across both single-image and multi-image tasks.

\item We propose \OURS, a simple yet effective strategy that permutes image orders and rephrases text to address multi-image coreference, while a rollout filtering mechanism improves training efficiency by focusing on informative samples.

\item As shown in Fig.~\ref{fig:multi_modal_benchmark}, extensive experiments on both in-domain and out-of-domain benchmarks demonstrate that our method achieves state-of-the-art performance on multi-image tasks and competitive results on single-image tasks. \end{itemize}

% Please read the instructions below carefully and follow them faithfully.

\section{Related Works}

\subsection{Multimodal Reasoning} 
Early efforts in multimodal reasoning \cite{mitra2024compositional,yao2024mulberry,li2025system} mainly focus on supervised fine-tuning (SFT) especially the dataset construction with chain-of-thought (CoT) reasoning steps. Recently, Deepseek-R1-Zero \cite{guo2025deepseek} demonstrates remarkable reasoning capabilities, spearheading large-scale reinforcement learning (RL) research in LLM. While VLMs also require substantial reasoning capabilities for vision-and-language tasks, recent investigations employ reinforcement learning to multimodal reasoning and present early progress including Visual-RFT \cite{liu2025visual}, LMM-R1 \cite{peng2025lmm}, VLM-R1 \cite{shen2025vlm}, Reason-RFT \cite{tan2025reason}, MM-EUREKA \cite{meng2025mm}, R1-OneVision \cite{yang2025r1}, Vision-R1 \cite{huang2025vision}, R1-V \cite{chen2025r1v}, R1-VL \cite{zhang2025r1}, VL-Rethinker \cite{wang2025vl}, VisualThinker-R1-Zero \cite{zhou2025r1}, OpenVLThinker \cite{deng2025openvlthinker} and Perception-R1 \cite{yu2025perception}. Most of these works adopt GRPO \cite{shao2024deepseekmath} for RL scaling through \emph{data} construction \cite{tan2025reason,huang2025vision,meng2025mm,zhou2025r1} or mixture \cite{peng2025lmm}, multi-stage \emph{training} strategy \cite{peng2025lmm,deng2025openvlthinker,huang2025vision,meng2025mm} or perception-wise \cite{liu2025visual,yu2025perception,shen2025vlm} or step-wise \cite{zhang2025r1} \emph{reward} design .
%text data mixing\cite{peng2025lmm}, multimodal CoT dataset construction\cite{tan2025reason,huang2025vision}, multimodal reasoning dataset\cite{meng2025mm,zhou2025r1}, multi-stage training\cite{peng2025lmm,shen2025vlm,deng2025openvlthinker,huang2025vision}, and reward function design \cite{liu2025visual,yu2025perception}. 
Distinct from these approaches, VL-Rethinker \cite{wang2025vl} strategically select high-value samples to avoid vanishing advantages, while NoisyRollout \cite{liu2025noisyrollout} introduces moderately distorted images as diverse samples for broad exploration to address the challenge of imperfect visual perception. 

\subsection{Interleaved VLM} With the significant advancement in single-image tasks, vision-language models (VLMs) \cite{2023GPT4VisionSC,team2023gemini,abdin2024phi,chen2024far,wu2024deepseek,bai2025qwen2,wu2024deepseek,li2024llava,chen2024internvl} have emerged, \emph{pretraining} on interleaved image-text datasets to address more comprehensive multi-image tasks in real-world scenarios.
Building on this, \cite{li2024llava,jiang2024mantis,li2025migician} further construct interleaved multi-image-text data to \emph{post-train} VLMs with instruction tuning.
These approaches either exploit learnable cross-attention modules with a Perceiver Resampler \cite{alayrac2022flamingo,awadalla2023openflamingo,li2023mimic},
or utilize a linear projection layer \cite{li2024llava1,sun2024generative} to integrate pretrained language models with visual encoders. Nonetheless, multi-image tasks still face further challenges beyond the perception problem in single-image tasks, including  complex spatial and positional/temporal reasoning.
To address the bias, \cite{tian2025identifying} proposes SoFt Attention (SoFA), a simple, training-free method to mitigate the position bias of VLMs when processing multiple images. Differently, we propose to scale the interleaved VLM with reinforcement learning to incentive general multimodal reasoning capability and further design a comprehensive rollout strategy for coreference resolution. 
\begin{figure}[tb]
% \vskip 0.2in
\begin{center}
\centerline{\includegraphics[width=1.0\linewidth]{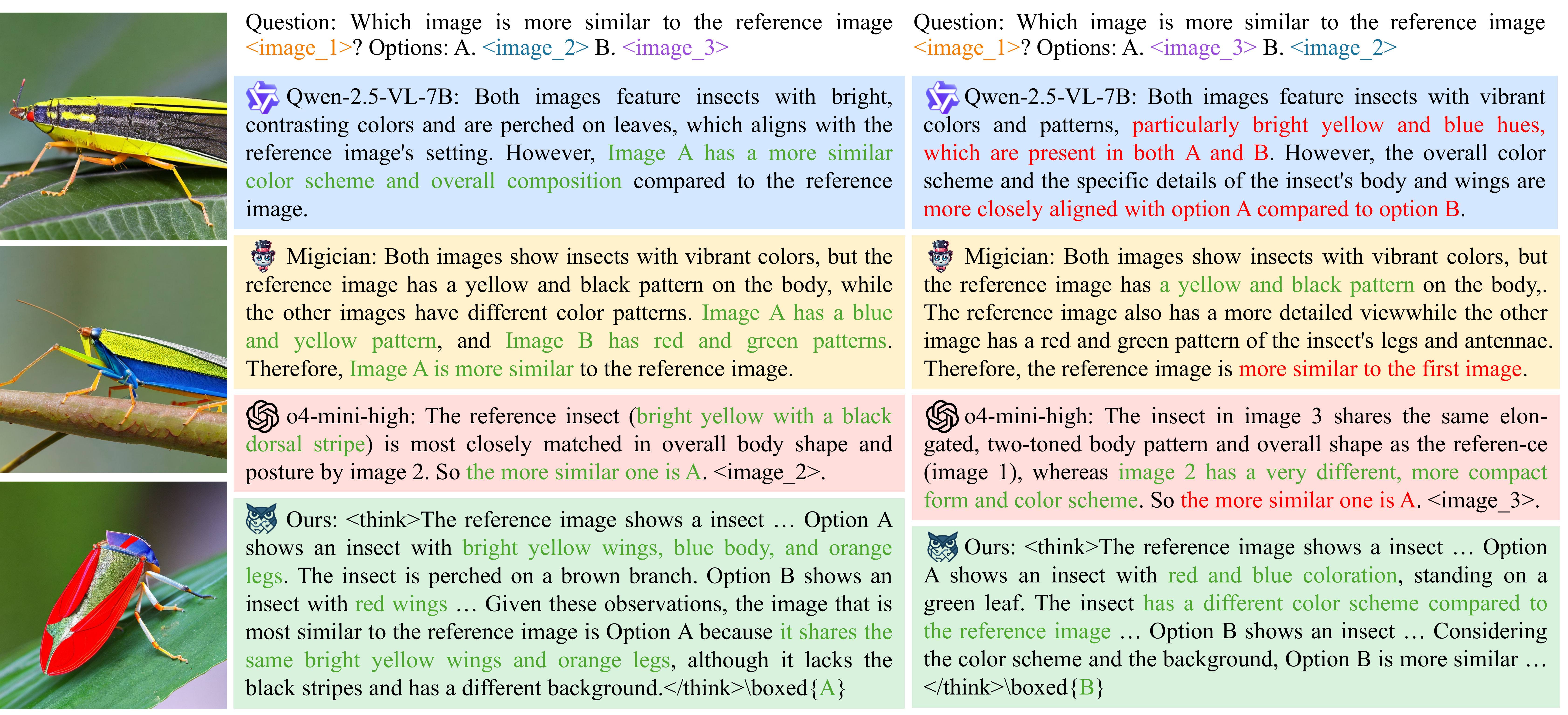}}
\caption{A showcase highlighting the challenges faced by existing VLMs: 
1) Fail to recognize or reason across multiple images; 
% 2) Struggle with co-referencing, failing to associate information with the correct image; 
2) VLMs exhibit positional bias, leading to inconsistent answers when the order of images is permuted even though the semantic of the input keeps the same.
   }
\label{fig:case}
\end{center}
\vskip -0.2in
\end{figure}

\section{Problem Formuation and Analysis}
The multimodal understanding task is defined as predicting an output answer given a text query interleaved with images. Formally, an input instance is represented as \( x = (I,Q) \), where \( Q \) denotes the textual query and \(I = \{i_j\}_{j=1}^n\) is the associated set of images. The corresponding output is denoted by \( y \), yielding a dataset \( D = \{(x_i, y_i)\}_{i=1}^N \). In practice, we interleave text tokens with image representations using a special placeholder token <image> to indicate the presence and position of an image within the input sequence. % \( I = \{i_1, i_2, \ldots, i_n\} \)
%The definition of multimodal understanding task is to predict the answer given the input text query interleaved with images. Formally, let $x=(Q,I)$ denote the input with both text query $Q$ and image set $I$, and $y$ represent the output%, and thus the dataset is $D=\{(x_{i},y_{i})\}$
%. In practice, we interleave text tokens with image representations using a special placeholder token <image> to indicate the presence and position of an image within the input sequence.

\subsection{Positional Bias}

%One of the critical challenge of multi-image multimodal reasoning is the positional correference which requires to understand the specific image corresponding to the given text. However, as also mentioned in \cite{tian2025identifying}, existing VLMs still suffer from performance drop significantly by the alteration of image orders. As shown in Fig. \ref{fig:case}, we summarized the three main problems of existing VLMs in handling multi-image tasks. Motivated by this, we propose to diversify the input images through permutation of the images and alter the text accordingly to guarantee the semantic invariance.

One of the critical challenges in multi-image multimodal reasoning is positional coreference, which requires accurately identifying the specific image that corresponds to a given textual reference. However, as highlighted in \cite{tian2025identifying}, existing vision-language models (VLMs) still suffer from significant performance degradation when the order of input images is altered. As illustrated in Fig. \ref{fig:case}, we summarize the major limitations of current VLMs in handling multi-image tasks. Motivated by these observations, we propose a simple yet effective strategy that diversifies image inputs through permutation, while simultaneously modifying the associated text to preserve semantic invariance. This approach encourages the model to develop a more robust understanding of positional relationships across images, thereby improving generalization in complex multimodal reasoning scenarios.

\subsection{Difficulty Imbalance}

Mantis-Instruct  \cite{jiang2024mantis} is the first multi-image instruction-tuning data, comprising 721K data. Therefore, we start with the Mantis-Instruct dataset as our training data. Before training, we firstly exploit the Qwen-2.5-VL \cite{bai2025qwen2} model to generate several outputs and calculate the average accuracy among these generations as the difficulty score of each sample for statistical analysis. Fig. \ref{fig:distribution} represents the difficulty distribution. We find that the distribution is extremely imbalanced, leading to inefficient and unstable reinforcement learning. Inspired by this finding, we further resample the data for a balanced difficulty distribution. In the following section, we discuss the pipeline to process the data according to the two observations.

\subsection{Data Prepocessing}
\begin{figure}[tb]
% \vskip 0.2in
\begin{center}
\subfloat[Multi-Stage Preprocessing Pipeline]{
  \includegraphics[width=0.48\linewidth]{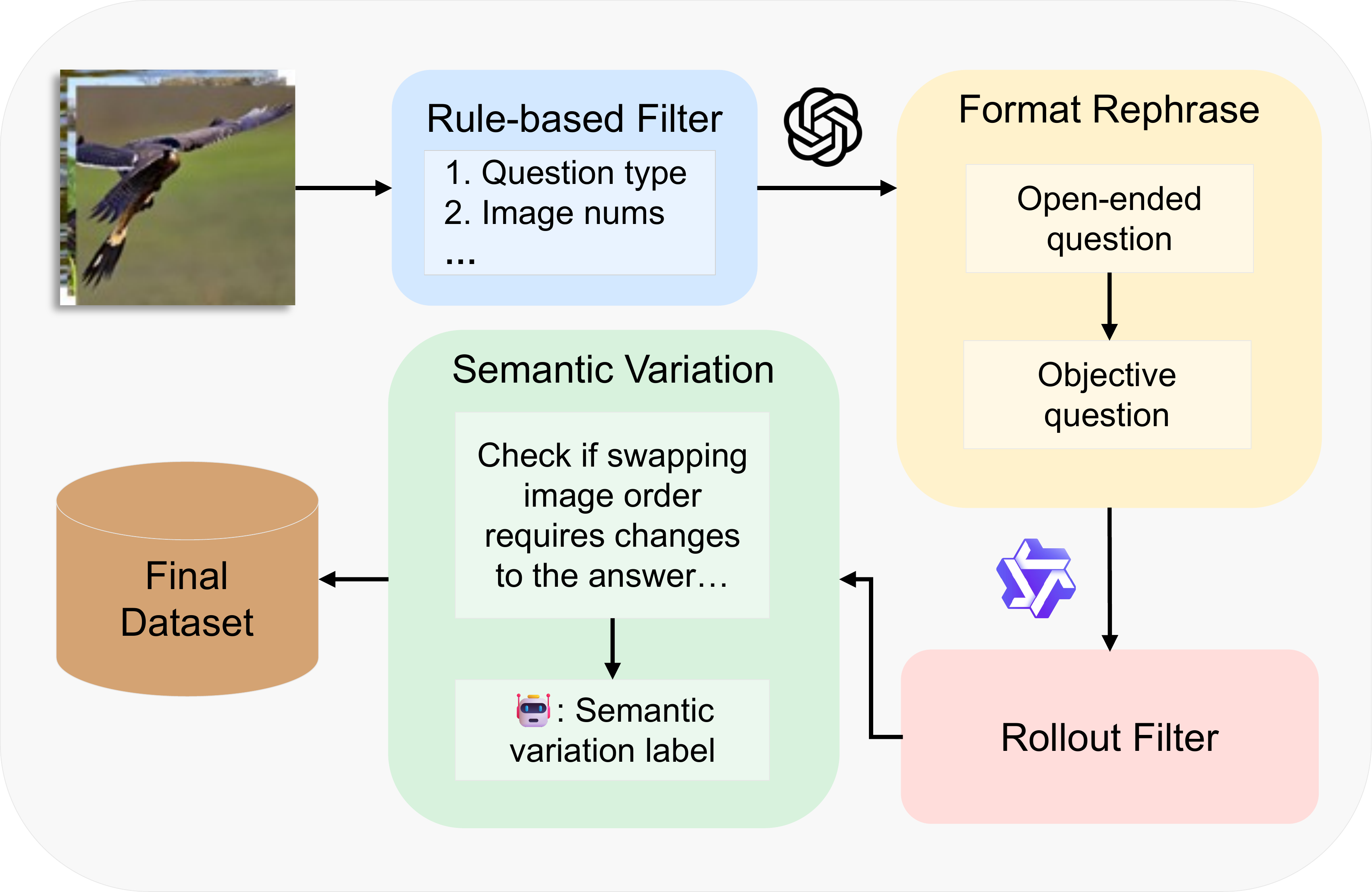}
  \label{fig:pipeline}
}
\hfill
\subfloat[Score Distribution Pre- and Post-Difficulty Filtering]{
  \includegraphics[width=0.48\linewidth]{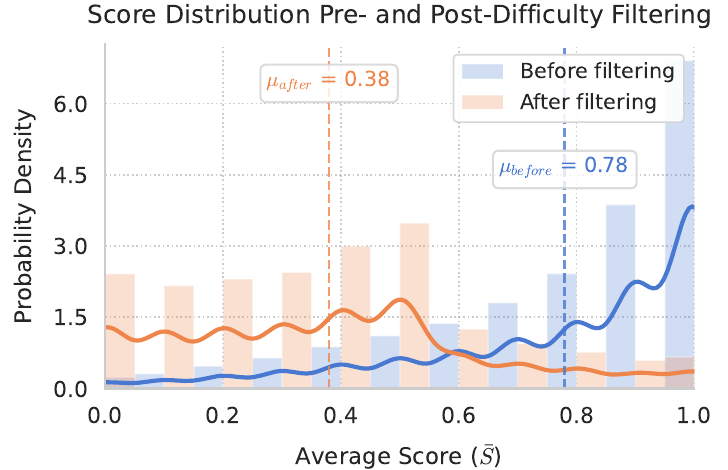}
  \label{fig:distribution}
}
\caption{Overview of Data Preprocessing and Difficulty Distribution.
(a) The multi-stage pipeline applies (i) rule-based filtering, (ii) format rephrasing, (iii) rollout-based filtering and (iv) semantic variation checking to curate QA pairs. 
(b) KDEs of the average score $\bar S$ before (blue) and after (orange) rollout-based difficulty filtering, with the mean falling from 0.78 to 0.38.}
\label{fig:data}
\end{center}
\vskip -0.2in
\end{figure}

We develop a multi-stage data processing pipeline that systematically filters and reformulates QA pairs from the original training corpus Mantis-Instruct~\cite{jiang2024mantis} to construct a higher-quality training subset, as illustrated in Fig.~\ref{fig:pipeline}.

Initially, we apply a rule-based filter to the original dataset, screening entries based on question types, image quantity, and other relevant attributes. 
Then we employ GPT-4o\cite{2023GPT4VisionSC} to perform format rephrasing on open-ended questions that are challenging to verify through rule-based methods. 
% This transformation converts ambiguous questions into verifiable formats, such as converting them into multiple-choice format, to better align with the requirements of GRPO training.
In the next phase, we implement a rollout filtering mechanism using Qwen2.5-VL-7B\cite{bai2025qwen2}. For each question, we perform rollout 10 times and calculate the average accuracy of the responses. Questions are then filtered based on their accuracy score, resulting in a refined distribution as shown in Fig.~\ref{fig:distribution}.
%This procedure enables us to identify and retain questions that demonstrate consistent performance patterns.

To evaluate the semantic sensitivity of questions to image order, we introduce a semantic variation analysis step using GPT-4o. Specifically, as illustrated in Fig.~\ref{fig:pipeline}, given an input–output pair \((x, y)\), %where \(x = (I, Q)\) consists of a set of images \(I = \{i_j\}_{j=1}^n\) and a textual question $Q$, 
we apply a random permutation \(\sigma\) to reorder the image set, resulting in a new input \(\hat{x} = (\hat{I}, Q)\) with \(\hat{I} = \{i_{\sigma(j)}\}\). GPT-4o is then prompted to assess whether the permuted input $\hat{x}$ preserves the semantics of the original input $x$, i.e., whether the intended answer should remain unchanged. The corresponding output \(\hat{y}\) is defined as:
\vspace{-1pt}
\begin{equation}
\label{eq:yhat}
\hat{y} = S(x, \hat{x}) \cdot y + (1 - S(x, \hat{x})) \cdot \Lambda(y, \sigma)
\end{equation}
\vspace{-1pt}
Here, \(S(x, \hat{x}) \in \{0,1\}\) is a semantic equivalence indicator obtained by prompting GPT-4o, where \(S = 1\) denotes the input semantics are preserved under permutation. The function \(\Lambda(y, \sigma)\) denotes the transformation applied to the answer when semantic equivalence does not hold. It adapts \(y\) to the permuted context according to the structure of the answer space \(Y\), such as reindexing visual choices.

This framework facilitates controlled perturbation of the visual input via image sequence permutations, aiming to simulate diverse spatial and temporal configurations and thereby encourage richer generalization and reasoning capabilities.

\section{Methods}
\label{headings}
\begin{figure}[tb]
% \vskip 0.2in  
\begin{center}
\centerline{\includegraphics[width=1.0\linewidth]{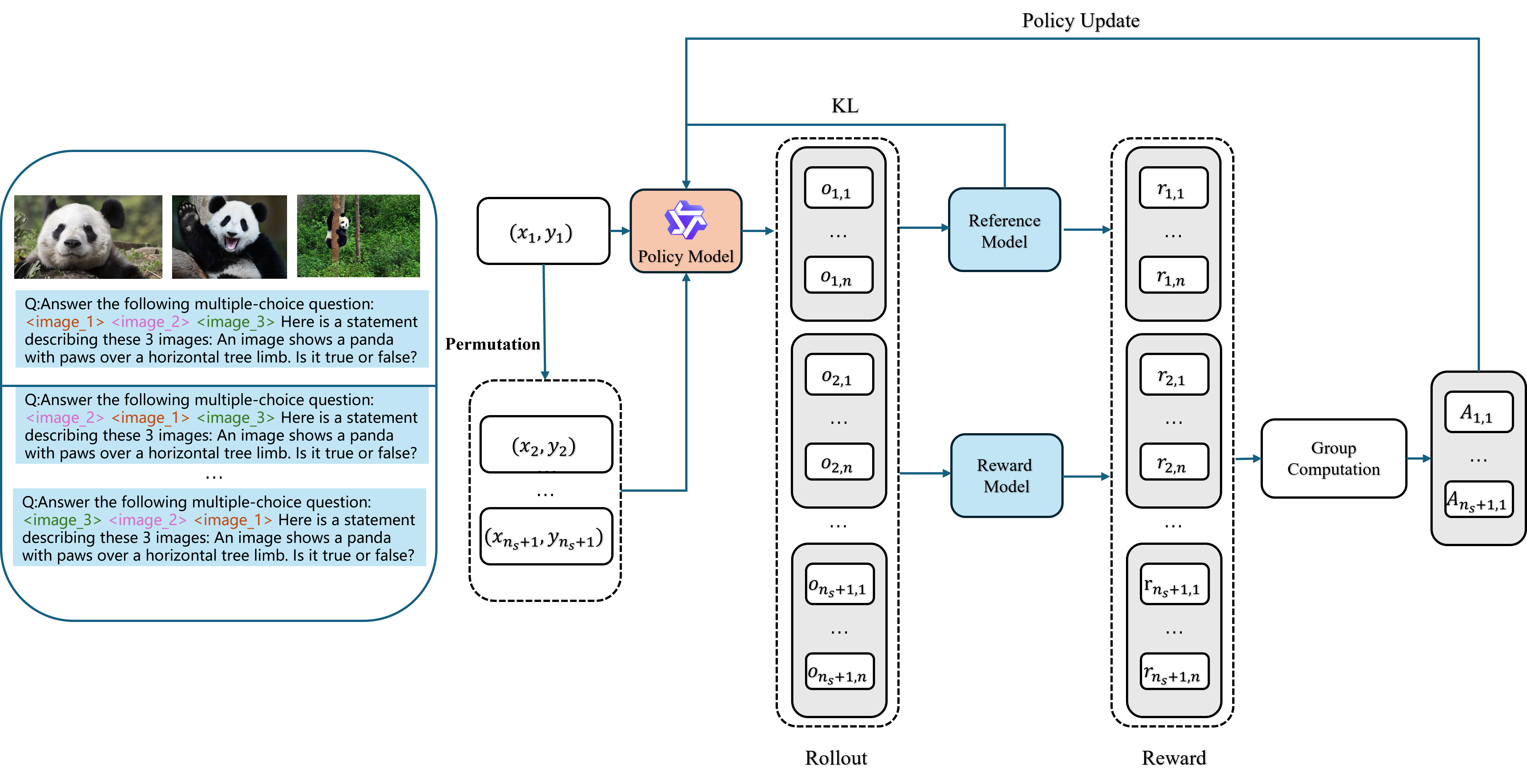}}
\caption{An illustration of \OURS~framework.   ($x_1,y_1$) denote original input sample, while $\{(x_i,y_i)\}_{i=2}^{n_s + 1}$ represent samples generated through permutation operation,  $n_s$ is the number of permutations per sample and $n$ is the number of rollouts per order. $o_{i,k}$,$r_{i,k}$ and $A_{i,k}$ represent the trajectory, reward and advantage of $k_{th}$ rollout of sample $(x_i, y_i)$ respectively.} 
\label{fig:training}
\end{center}
\vskip -0.2in
\end{figure}

%\subsection{Preliminary}
The goal of multimodal model is to train a policy model $\pi_{\theta}(y|x)$ parameterized by $\theta$ to optimize the correctness of model response $o$. In this paper, we exploit reinforcement learning (RL) framework specifically GRPO to learn to improve the multimodal reasoning as presented in Fig.\ref{fig:training}. 
Firstly, the policy model $\pi_{\theta}(y|x)$ takes the input $x$ to generate a group of responses $\{o_{i}\}$. Next, the reward model calculates the score $r(x,o_{i},y)$ for each response $o_{i}$, which adopts rule-based functions to evaluate the accuracy and format correctness. Furthermore, the group computation module estimates the advantage $A_{i}$ for each response $o_{i}$ to update the policy model. 

Specifically, we propose a plug-and-play module to improve the efficient and effective training through diversified advantage estimation and dynamic resampling.
1) Before exploration, we resample rollouts with variable reward distribution to exploit learned policies effectively. This ensures the learning focuses on trajectories that contribute most to learning optimal behaviors across both spatial and positional dimensions.2) To mitigate positional bias in interleaved multi-image reasoning, we introduce diversity in the rollout phase:
for Single-image,we keep the original format.
for multi-image positional diversity, use permutations of image sequences to simulate varied positional relationships.  

\subsection{GroupRelative Policy Optimization}
GRPO \cite{shao2024deepseekmath} estimates the advantage score based on the reward normalization of a group of responses given a specific query. Particularly, the advantage score $A_{i}$ and the corresponding loss are defined as: 
\begin{equation}
A_{i}=\frac{r(x,o_{i},y)-\frac{1}{|o|}\sum_{j=1}^{|o|}r(x,o_{j},y)}{\sqrt{\frac{1}{|o|}\sum_{j=1}^{|o|}\left(r(x,o_{j},y)-\frac{1}{|o|}\sum_{j=1}^{|o|}r(x,o_{j},y)\right)^2}} 
\end{equation}
%, which expel the value model in PPO \cite{schulman2017proximal}.
% \begin{equation}
% \mathcal{L}_{GRPO} = \frac{1}{|o|}\sum_{i=1}^{|o|}\left(\frac{1}{|o_{i}|} \sum_{t=1}^{|o_{i}|} min [\frac{\pi_{\theta}(o_{i,t}|x,o_{i,<t})}{\pi_{old}(o_{i,t}|x,o_{i,<t})}A_{i},clip(\frac{\pi_{\theta}(o_{i,t}|x,o_{i,<t})}{\pi_{old}(o_{i,t}|x,o_{i,<t})},1-\epsilon,1+\epsilon)A_{i}] 
% \\
% - \beta
% \mathcal{D}_{KL}(\pi_{\theta}||\pi_{ref})\right)
% \end{equation}
\vspace{-4pt}
\begin{equation}
\mathcal{L}_{GRPO} = \frac{1}{|o|}\sum_{i=1}^{|o|}\left(  \min \big(\frac{\pi_{\theta}(o_i|x)}{\pi_{old}(o_i|x)}A_{i},\text{clip}(\frac{\pi_{\theta}(o_i|x)}{\pi_{old}(o_i|x)},1-\epsilon,1+\epsilon)A_{i}\big) \\
- \beta \mathcal{D}_{KL}(\pi_{\theta}||\pi_{ref})\right)
\end{equation}
\vspace{-2pt}
where $|o|$ is the group size, $o_{i}$ is the $i_{th}$ response given the input $x$. $\pi_{old}$ is the old policy model to generate the response $o$, and $\pi_{\theta}$ is the current policy model to optimize. $r(x,o_{i},y)$ is the rule-based reward function to calculate the accuracy and format correctness of the response $o_{i}$.

\subsection{Permutation GRPO}

Let $(x_1,y_1)$ denote the original input sample, where $x_1$ consists of a query and one or more images, and $y_1$ is the corresponding answer. We augment it in to a group of samples $\{(x_2,y_2),(x_3,y_3)...(x_{n_s+1},y_{n_s+1})\}$.  For multi-image input,
we augment it by swapping images in $x$ randomly and modify the the answer $y$ correspondingly. To ensure training stability, we introduce a linearly decaying factor $\alpha_t$ (scheduled over training steps $t$) that controls the probability of applying image swapping per training batch. In the rollout progress, we input the diversified samples to the policy model to generate response $o_i$ and  the reward model for further reward calculation. After that, we merge all $n_s+1$ groups of responses together to calculate the baseline, final advantage and loss function as:
\vspace{-3pt}
\begin{equation}
    \label{eq.baseline}
    \bar R = \frac{1}{\sum_{i=1}^{n_s+1} |o_i|}\sum_{i=1}^{n_s+1}\sum_{k=1}^{|o_i|} r(x_i,o_{i,k},y_i)
\end{equation}
\begin{equation}
    \label{eq.adv}
    A_{i,k}=\frac{r(x_i,o_{i,k},y_i)-\bar R}{\sqrt{\frac{1}{\sum_{i=1}^{n_s+1}|o_i|}\sum_{i=1}^{n_s+1} \sum_{k=1}^{|o_i|}(r(x_i,o_{i,k},y_i)-\bar R)^2}}
\end{equation}
% 需要改成实际使用的 token level grpo
\begin{equation}
\label{eq.loss}
\begin{split}
\mathcal{L}(\theta) &= \frac{1}{\sum_{i=1}^{n_s+1}|o_i|} \Bigg( 
\sum_{i=1}^{n_s+1}\sum_{k=1}^{|o_i|} \min \left( 
\frac{\pi_{\theta}(o_{i,k} \mid x_i)}{\pi_{\text{old}}(o_{i,k} \mid x_i)} A_{i,k},\ 
\text{clip}\big( 
\frac{\pi_{\theta}(o_{i,k} \mid x_i)}{\pi_{\text{old}}(o_{i,k} \mid x_i)},\ 
1 - \epsilon,\ 
1 + \epsilon 
\big) A_{i,k} 
\right) \\
&\quad - \beta \mathcal{D}_{\mathrm{KL}} \left( \pi_{\theta} \,\|\, \pi_{\text{ref}} \right) 
\Bigg)
\end{split}
\end{equation}
\vspace{-2pt}
where $o_{i,k}$ denote the $k_{th}$ response for input $x_i$, and $A_{i,k}$ represent the advantage of this rollout. In our experiments, we set the number of responses per order $|o_i|$ to a same value, i.e. $|o_1|=|o_2|=\cdots=|o_{n_s+1}|=n$. Thus the actual group size is $\sum_{i=1}^{n_s+1}|o_i|$.

% why swap multi_modal_inputs?
Note that unlike NoisyRollout\cite{liu2025noisyrollout}, which performs policy update solely conditioned on the original images, we calculate policy loss of each rollout group $o_i$ based on augmented inputs $x_i$, as shown in Equation \ref{eq.loss}. This is because the answer is also updated after permutation, continuing to use the original prompt as a condition is improper and might lead to model collapse.
% 1. Since the answer updates after image swapping, continuing to use the original prompt as a condition is incorrect and might lead to model collapse

% \subsection{Dynamic Resampling/Online Filtering} 
% According to our analysis, we discover the divergence and imbalance of the distribution of the advantage score, which makes the training unstable and inefficient. Similarly, Decouple Clip and Dynamic sAmpling Policy Optimization (DAPO) \cite{yu2025dapo} also claimed the existing RL algorithm suffer from advantage vanishing for efficient and effective training. To mitigate gradient-decreasing problem, DAPO proposed to over-sample and filter out prompts with the accuracy equal to 1 and 0. In this paper, We take a step further and propose to dynamically sample by smoothing the advantage score distribution. %for an efficient and effective training. 

% \subsection{Annealing Schedule}

% Annealing only train without the diversified advantage.

\begin{algorithm}[htb]
\caption{\OURS: Permutation-Enhanced Reinforcement Learning}
\label{alg:token_alignment}
\begin{algorithmic}
   % \STATE \textbf{Input:}
   % %\STATE \quad Student tokens \(\{x_i^S\}\) with offsets \(\{[st_i^S, ed_i^S]\}\)
   % %\STATE \quad Teacher tokens \(\{x_j^T\}\) with offsets \(\{[st_j^T, ed_j^T]\}\)
   % \STATE \textbf{Output:} %Mapping \(\mathrm{mapping}[i]\)

   % \FOR{each \(i\)-th student token \(x_i^S\)}
   %     \STATE (1) {Get student character-level offsets}:
   %     \STATE \quad \quad \(st_i^S, ed_i^S \leftarrow x_i^S\)

   %     % \STATE

   %     \STATE (2) {Find overlapping teacher range via binary searches:}
   % \ENDFOR

   % \STATE \textbf{return} \(\mathrm{mapping}\)
\STATE \textbf{Input:} Current policy $\pi_{\theta}$, old policy $\pi_{\theta_{old}}$, dataset $\mathcal{D}$, training steps $T_{\text{max}}$, clip parameter $\epsilon$, initial decay factor $\alpha_0$, linear decay scheduler $\eta(\cdot)$, permutation operation $P(\cdot )$, permutation number $n_s$, rollout number per order $n$
\FOR{$t = 1$ to $T_{\text{max}}$}
    \STATE Sample batch $(\mathbf{x_1},\mathbf{y_1}) \sim \mathcal{D}$, $\mathcal{B}=\{(\mathbf{x_1},\mathbf{y_1})\}$
    \STATE Set decay factor $\alpha_t = \eta(\alpha_0, t, T_{\text{max}})$ 
    % \COMMENT{Annealing schedule}
    \FOR{$i=2$ to $n_s+1$}
    \STATE Generate randomly swapped images and modify answers correspondingly, $(\mathbf{x_i},\mathbf{y_i}) = P((\mathbf{x_1},\mathbf{y_1}), \alpha_t)$
    \STATE $\mathcal{B} \leftarrow \mathcal{B} \cup (\mathbf{x_i},\mathbf{y_i})$
    \ENDFOR
    \FOR{each sample $(\mathbf{x}_i,\mathbf{y}_i)$ in $\mathcal{B}$}
        \STATE Generate responses $\{\mathbf{o}_{i,k}\}_{k=1}^{n}$ from $\pi_{\theta}(\mathbf{o} \mid \mathbf{x}_i)$ 
        % \STATE Compute rewards $R_{i,k} = r(\mathbf{x}_i, \mathbf{o}_{i,k}, \mathbf{y}_i)$ for $k \in \{1, \ldots, n\}$
        
    \ENDFOR
    % \STATE Sample $\{\mathbf{o}_j\}_{j=1}^{n_1}$ from $\pi_{\theta_{\text{old}}}(\mathbf{o} \mid I, \mathbf{q})$ 
    
    % \STATE Sample $\{\mathbf{o}_j\}_{j=1}^{n_1}$ from $\pi_{\theta_{\text{old}}}(\mathbf{o} \mid I, \mathbf{q})$ 
    % \COMMENT{Clean rollouts}
    % \STATE Sample $\{\mathbf{o}_k\}_{k=n_1+1}^{n_1+n_2}$ from $\pi_{\theta_{\text{old}}}(\mathbf{o} \mid \tilde{I}, \mathbf{q})$ 
    % \COMMENT{Noisy rollouts}
    
    % \STATE Compute rewards $R_i = R(I, \mathbf{q}, \mathbf{o}_i)$ for all $i \in \{1, \ldots, n_1+n_2\}$
    \STATE Compute baseline $\bar{R}$ according to Eq.\ref{eq.baseline}
    \STATE Compute advantages $\hat{A}_i$ according to Eq.\ref{eq.adv}
    
    \STATE Update policy according to Eq.\ref{eq.loss}
    % \STATE $L(\theta) = \mathbb{E}\Big[ \frac{1}{n_1+n_2}\sum_{i=1}^{n_1+n_2} \min\Big(\frac{\pi_{\theta}(\mathbf{o}_i \mid I, \mathbf{q})}{\pi_{\theta_{\text{old}}}(\mathbf{o}_i \mid I, \mathbf{q})}\hat{A}_i, \mathrm{clip}\big(\frac{\pi_{\theta}(\mathbf{o}_i \mid I, \mathbf{q})}{\pi_{\theta_{\text{old}}}(\mathbf{o}_i \mid I, \mathbf{q})}, 1-\epsilon, 1+\epsilon\big)\hat{A}_i \Big) \Big]$
    
    \STATE $\theta \leftarrow \theta - \nabla_{\theta}\mathcal{L}(\theta)$, $\theta_{old} \leftarrow \theta$
    % \COMMENT{Update conditioned on clean images only}
    % \STATE $\theta_{\text{old}} \leftarrow \theta$ \COMMENT{Update old policy parameters}
\ENDFOR
\end{algorithmic}
\end{algorithm}

\section{Experiment}
\paragraph{Training}  
Our training data comprise two parts: 22K multi-image instruction examples curated from the 721K examples in Mantis-Instruct~\cite{jiang2024mantis}, and 36K single-image examples from the K12 dataset for RL. We initialize our policy with Qwen2.5-VL-7B-Instruct~\cite{bai2025qwen2} and build on the veRL framework~\cite{sheng2024hybridflow}. During RL fine-tuning, we apply one random permutation per sample ($n_s=1$) and generate six responses per order ($n=6$), yielding 12 rollouts per input. We set the KL coefficient $\beta=0.01$, train for 2 epochs with a learning rate of $1\times10^{-6}$ and a batch size of 256. Further details are provided in the appendix.

% Our training datasets contains multi-image datasets and single-image datasets separately or together. There are 721K multi-image instruction data in Mantis\cite{jiang2024mantis}, from which we cooked about 22k dataset for training. Besides, We use 36k k12 single-image dataset for RL training.  %Projudge Process Judges 
% We adopt Qwen2.5-VL-7B-Instruct \cite{bai2025qwen2} as our initialized policy model. We construct our model on top of the general framework veRL  \cite{sheng2024hybridflow}. 
% % setting of training: Full or lora, vision encoder is frozen or not, parameters: rollout counts, $\beta$ is set to... GPU....system prompt? 
% For each sample, we conduct 1 permutation ($n_s=1$) and generate 6 responses ($n=6$) per order, resulting in 12 rollouts per input. We set $\beta$ to 0.01 and train for 2 epochs with learning rate of 1e-6 and batch size of 256. 
 
\paragraph{Benchmarks} We conduct experiments on both multi-image benchmarks and single-image benchmarks. As the main experiment, we employ Mantis-Eval\cite{jiang2024mantis}, BLINK\cite{fu2024blink}, %MuirBench\cite{wang2024muirbench}, 
MMIU\cite{meng2024mmiu} %and MIRB\cite{zhao2024benchmarking} 
as multi-image benchmarks. Furthermore, we evaluate the generalization on widely used single-image benchmarks including MathVista\cite{lu2023mathvista}, MathVerse\cite{zhang2024mathverse} and MathVision\cite{wang2024measuring}. Besides, we also evaluate our model on out-of-domain multi-image benchmarks including Remi \cite{kazemi2024remi} and MV-Math \cite{wang2025mv}. We directly employ the VLMEvalKit \cite{duan2024vlmevalkit} to evaluate the performance.

%Here, we list some of the selected tasks: forensic detection and imagejigsaw from BLINK[15], difference spotting and  diagram understanding from MuirBench [56], visual analogy and attribute matching from MIRB [66]. We include all examples from Mantis-Eval [19] since the majority are position-agnostic. 

\subsection{Main Results}
% \paragraph{Baselines} We compare \OURS~with three categories of baseline models: \emph{Proprietary models} including GPT-4o\cite{2023GPT4VisionSC}; \emph{General Models} like VILA-1.5, Mantis-Idefics2, LLaVA-v1.5-7B, LLaVA-NeXT-Interleave\cite{li2024llava}, LLaVA-OneVision\cite{li2024llava1}, LLaVA-OneVision-+VISC-150K, Migician-630K\cite{li2025migician}, InternVL2\cite{chen2024internvl}, InternVL2.5, Deepseek-VL-7B\cite{wu2024deepseek}, Qwen2-VL-7B, Qwen2.5-VL-7B\cite{bai2025qwen2}; \emph{Reasoning models} containing R1-VL-7B-260K\cite{zhang2025r1}, Qwen2-VL-7B-VISC-150K \cite{wang2025vl}, R1-Onevision-7B\cite{yang2025r1}, MM-Eureka\cite{meng2025mm}, Noisy-K12\cite{liu2025noisyrollout}.
\paragraph{Baselines} The compared models adopt different training strategies. LLaVA-v1.5-7B\cite{liu2024improved}, LLaVA-NeXT-Interleave\cite{li2024llava}, LLaVA-OneVision\cite{li2024llava1}, Qwen2-VL-7B\cite{wang2024qwen2}, Qwen2.5-VL-7B\cite{bai2025qwen2}, Deepseek-VL-7B\cite{wu2024deepseek}, InternVL2\cite{chen2024internvl}, and InternVL2.5\cite{chen2024expanding} are instruction-tuned base models. Migician-630K\cite{li2025migician} is further fine-tuned on MGrounding-630K to enhance multi-image grounding. Models like LLaVA-OneVision-+VISC-150K\cite{zhang2025weaving} and Qwen2-VL-7B-VISC-150K\cite{wang2025vl} are fine-tuned on synthetic VISC-150K dataset, which explicitly encodes inter-image relations.

R1-VL-7B-260K\cite{zhang2025r1} and R1-OneVision-7B\cite{yang2025r1} are trained on 260K and 155K supervised samples before RL. MM-Eureka\cite{meng2025mm} and Noisy-K12\cite{liu2025noisyrollout} are trained only with RL on K12 single-image math data. Our method follows the same RL-only setting, but focuses on multi-image scenario.
\paragraph{SOTA Results}
\definecolor{multiimg}{RGB}{255,200,200}
\definecolor{singlemath}{RGB}{180,210,255}
\definecolor{multiimgmath}{RGB}{190,255,190}

\begin{table}[tb]
\centering
\small
\caption{
Model performance across VQA and math reasoning benchmarks. 
Among Instruct‑ and RL‑tuning models, \textbf{bold} numbers denote the best value in each column, 
and \underline{underlined} numbers are the second‑best. 
% \textcolor{multiimg}{Multi‑image Natural‑scene Understanding}; 
% \textcolor{singlemath}{Single‑image Math Reasoning}; 
% \textcolor{multiimgmath}{Multi‑image Math Reasoning}. 
Task categories: \textcolor{multiimg}{multi-image natural scene understanding}, \textcolor{singlemath}{single-image math reasoning}, and \textcolor{multiimgmath}{multi-image math reasoning}.
The last column (AVG) is the mean of all eight metrics for rows with complete scores. Models marked with \textsuperscript{*} are evaluated via vLLM inference.
}

\setlength{\tabcolsep}{5pt}
\renewcommand{\arraystretch}{1.2}

\resizebox{\linewidth}{!}{
\begin{tabular}{
l l
>{\columncolor{multiimg}}c
>{\columncolor{multiimg}}c
>{\columncolor{multiimg}}c
>{\columncolor{singlemath}}c
>{\columncolor{singlemath}}c
>{\columncolor{singlemath}}c
>{\columncolor{multiimgmath}}c
>{\columncolor{multiimgmath}}c
c
}
\toprule
\textbf{Model} & \textbf{Size} 
& \textbf{Mantis‑Eval} & \textbf{BLINK} & \textbf{MMIU} 
& \textbf{MathVista} & \textbf{MathVerse} & \textbf{MathVision}
& \textbf{Remi} & \textbf{MV‑MATH} & \textbf{AVG} \\
\midrule
Human & -- & -- & 95.67 & -- & 60.30 & 64.90 & 68.82 & 95.80 & 76.50 & --\\
Random & -- & -- & 38.10 & 27.40 & 17.90 & 12.40 & 7.17 & -- & -- & --\\
GPT‑4o & -- & 68.00 & 68.00 & 55.70 & 63.80 & 37.60 & 30.60 & 59.96 & 32.10 & --\\
\midrule
\multicolumn{11}{l}{%Instruct‑tuning models
\textit{Multimodal General Models}} \\
VILA‑1.5 & 8B & 51.15 & 39.30 & 32.45 & 65.40 & -- & -- & -- & -- & --\\
Mantis‑Idefics2 & 8B & 57.14 & 49.05 & 45.60 & 40.40 & 15.05 & 12.40 & -- & 5.50 & --\\
LLaVA‑v1.5‑7B & 7B & 31.34 & 37.13 & 19.20 & 24.20 & 14.82 & 11.74 & -- & 15.50 & --\\
LLaVA‑NeXT-Interleave & 7B & -- & 51.30 & 32.40 & 34.00 & 17.99 & 13.81 & -- & 14.70 & --\\
LLaVA‑OneVision & 7B & 64.20 & 48.20 & 40.32 & 63.20 & 26.20 & 18.30 & 13.69 & 18.80 & 36.61\\
LLaVA‑OneVision-VISC‑150K & 7B & 66.36 & 50.24 & 46.52 & -- & -- & -- & -- & -- & --\\
Migician‑630K & 7B & 69.12 & 51.53 & 52.32 & 58.90 & 30.10 & 19.60 & 8.73 & 14.88 & 38.15\\
InternVL2 & 8B & 65.40 & 50.90 & 42.00 & 58.30 & 37.00 & 17.40 & 25.85 & -- & --\\
InternVL2.5 & 8B & 67.00 & 54.80 & 46.70 & 64.40 & 39.50 & 19.70 & 30.92 & 16.84 & 42.48\\
Deepseek‑VL‑7B & 7B & -- & 40.90 & 24.64 & 37.20 & 18.42 & 13.82 & -- & 14.50 & --\\
Qwen2‑VL‑7B & 7B & 69.60 & 53.20 & \underline{52.76} & 58.20 & 32.50 & 16.30 & 28.62 & 14.53 & 40.71\\
Qwen2.5‑VL‑7B & 7B & 70.80 & 55.23 & 52.00 & 68.20 & 46.30 & 25.07 & 36.38 & 20.41 & 47.90\\
\midrule
\multicolumn{11}{l}{ 
\textit{Multimodal Reasoning Models}} \\
R1‑VL‑7B‑260K\textsuperscript{*} & 7B & 64.81 & 55.00 & 46.58 & 63.50 & 40.00 & 24.70 & 24.31 & 18.77 & 42.62\\
Qwen2‑VL‑7B‑VISC‑150K\textsuperscript{*} & 7B & 69.12 & 55.34 & \underline{52.76} & 58.50 & -- & -- & -- & -- & --\\
R1‑Onevision‑7B\textsuperscript{*} & 7B & 71.76 & 32.38 & 46.48 & 64.10 & 46.40 & \textbf{29.90} & 36.96 & 19.06 & 43.38\\
MM‑Eureka\textsuperscript{*} & 7B & 69.91 & 51.49 & 51.67 & \textbf{73.00} & \underline{50.30} & 26.90 & \underline{43.31} & 24.39 & 49.49\\
Noisy‑K12\textsuperscript{*} & 7B & \underline{73.15} & 54.23 & 52.13 & \underline{72.90} & \textbf{52.80} & \underline{28.90} & 41.27 & \underline{24.99} & \underline{50.05}\\
% \textbf{InternVL-PeRL(Ours)\textsuperscript{*}}&7B &70.1 & \underline{56.40} & 46.49 & 67.3& 40.2 & 23.1 & 34.23 &20.05& 44.73\\
\textbf{Qwen-PeRL (Ours)\textsuperscript{*}} & 7B & \textbf{76.39} & \textbf{58.53} & \textbf{54.23} & \textbf{73.00} & 49.56 & 28.26 & \textbf{43.38} & \textbf{25.68} & \textbf{51.13}\\
$\Delta$ \textit{over the Baseline Model} & -- 
& \textcolor{red}{+5.59} & \textcolor{red}{+3.30} & \textcolor{red}{+2.23} 
& \textcolor{red}{+4.80} & \textcolor{red}{+3.26} & \textcolor{red}{+3.19} 
& \textcolor{red}{+7.00} & \textcolor{red}{+5.27} & \textcolor{red}{+3.23} \\
\bottomrule
\end{tabular}}
\label{tab:vqa_math}
\end{table}

As shown in Table~\ref{tab:vqa_math}, PeRL demonstrates exceptional performance across diverse benchmarks, achieving state-of-the-art results on multimodal natural scene understanding while maintaining competitive performance on mathematical reasoning tasks.

Our model achieves superior results with 76.39 on Mantis-Eval and 58.53 on BLINK, substantially outperforming previous best models. This achievement is particularly significant considering our approach utilizes substantially less training data compared to traditional supervised fine-tuning methods. On single-image mathematical reasoning tasks, PeRL demonstrates competitive performance with scores of 73.00 on MathVista and 49.56 on MathVerse, comparable to specialized state-of-the-art models MM-Eureka and Noisy-K12. 

Most notably, our model exhibits robust generalization on multi-image mathematical reasoning tasks despite these benchmarks being out-of-distribution relative to our training paradigm. This generalization capability is particularly valuable as our model was only trained on single-image mathematical reasoning and multi-image natural scene understanding datasets. 

The comprehensive evaluation across 8 metrics yields an average score of 51.13, highlighting the effectiveness of our proposed training framework tailored specifically for interleaved multimodal tasks.

\subsection{Ablation Study}

\begin{figure}[tb]
% \vskip 0.2in
\centering
\begin{subfigure}[t]{0.48\linewidth}
    \centering
    \includegraphics[width=\linewidth]{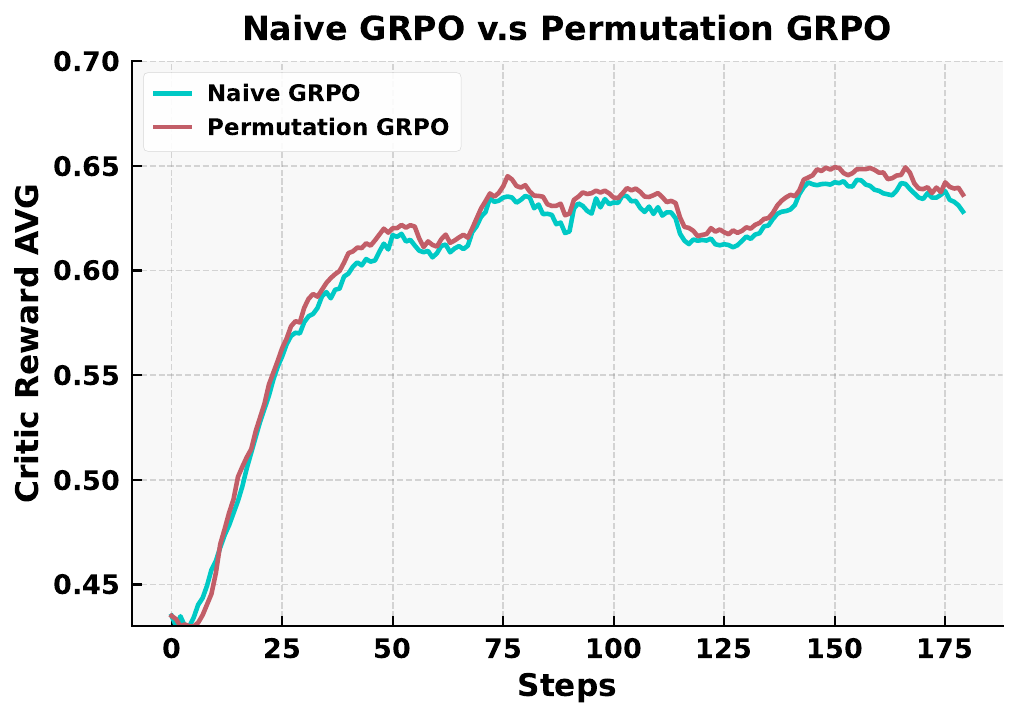}
    \caption{Reward curve comparison.}
    \label{fig:ablation_permutation_a}
\end{subfigure}
\hfill
\begin{subfigure}[t]{0.48\linewidth}
    \centering
    \includegraphics[width=\linewidth]{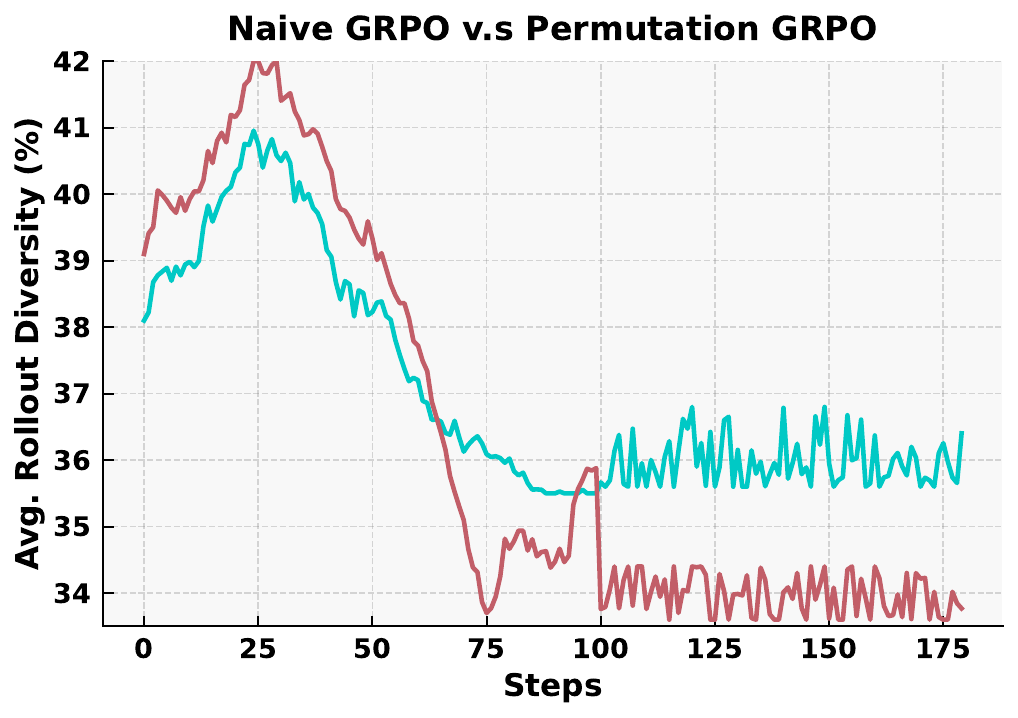}
    \caption{Rollout diversity comparison.}
    \label{fig:ablation_permutation_b}
\end{subfigure}
\caption{Permutation GRPO achieves more consistent training behavior.}
\label{fig:ablation_permutation}
\vskip -0.2in
\end{figure}

\paragraph{Permutation vs. Naive GRPO}
We investigate the effect of diversified permutation on training dynamics and generalization. As shown in Fig.~\ref{fig:ablation_permutation_a}, Permutation GRPO gradually surpasses naive GRPO with more stable training. This is because the permutation disrupts fixed visual token positions, exposing the model to harder cases initially, but enabling it to overcome positional bias through adaptation. 

To evaluate the impact on output diversity, we compute the average intra-batch cosine similarity of rollout embeddings using the BGE-Large model\footnote{\url{https://huggingface.co/BAAI/bge-large-en-v1.5}}. As shown in Fig.~\ref{fig:ablation_permutation_b}, Permutation GRPO maintains higher diversity throughout training. Given the same prompt, the policy sees varied visual token orders, which encourages learning order-invariant representations and improves robustness. 

% These improvements are also reflected in the evaluation performance shown in Table~\ref{table:permutation_ablation}, particularly on benchmarks requiring multi-image spatial reasoning.
% We analyze every comp
% We further quantify semantic diversity by encoding the textual rollouts using the embedding model bge-large-en-v1.5\footnote{\url{https://huggingface.co/BAAI/bge-large-en-v1.5}}\cite{xiao2024c}. Within each batch, we compute the average pairwise cosine similarity to assess the similarity among generated outputs. A lower cosine similarity indicates greater diversity. Results show that the permutation-based strategy consistently maintains a higher level of diversity across training steps. In contrast, the original GRPO baseline exhibits a sharp decline in diversity after a certain number of steps, suggesting early convergence and reduced output variation. These findings demonstrate that Diversified Permutation mitigates mode collapse and encourages the model to explore a wider range of output behaviors during training.

\paragraph{Permutation intensity}
We investigate the effect of the number of augmented permutations $n_s$. As shown in Table~\ref{table:permutation_ablation}, increasing $n_s$ from 0 (i.e., the naive GRPO baseline without permutation) to 2 improves performance on most benchmarks. For instance, Mantis-Eval increases from 75.93 to 77.78, and BLINK from 56.91 to 59.50, indicating that exposure to varied input orders promotes positional invariance. The improvements are particularly notable on multi-image tasks with complex spatial structures, such as BLINK and MathVerse. However, $n_s = 2$ yields marginal degradation on reasoning-centric benchmarks (e.g., MathVista and Remi) and incurs additional training cost due to reduced key-value cache reuse. These findings suggest a trade-off: permutation augmentation enhances spatial generalization but may impair reasoning performance and training efficiency.

\begin{table}[tb]
  \centering
  \small
  \caption{Permutation Ablation: $n_s$ is the number of permuted samples per original input. The total number of rollouts per input $12$ is held constant. $n_s=0$ indicates no permutation applied, while $n_s=2$ indicates two permuted samples per input.}
  \label{table:permutation_ablation}
  \resizebox{\linewidth}{!}{%
  \begin{tabular}{lcccccccc}
    \toprule
    Permutation & \textbf{Mantis-Eval} & \textbf{BLINK} & \textbf{MMIU} & \textbf{MathVista} & \textbf{MathVerse} & \textbf{MathVision} & \textbf{Remi} & \textbf{MV-MATH} \\
    \midrule
    $n_s=0$  & 75.93 & 56.91 & 52.26 & 71.8  & 49.20 & 26.51 & 39.62 & 24.02 \\
    $n_s=1$ & 76.39 & 58.53 & 54.23 & 73.0  & 49.56 & 28.26 & 41.65 & 25.68 \\
    $n_s=2$ & 77.78 & 59.50 & 54.55 & 72.2  & 50.38 & 27.89 & 39.04 & 24.54 \\
    $n_s=3$ & 74.07 & 57.02 & 52.17 & 70.9  & 46.39 & 27.00 & 38.62 & 23.25 \\
    \bottomrule
  \end{tabular}
  }
\end{table}

\begin{table}[!tb]
  \centering
  \small
  \caption{Single v.s. Multiple Image Disentanglement Ablation}
  \label{table:single_multi_combined_ablation}
  \resizebox{\linewidth}{!}{%
  \begin{tabular}{cc|*{8}{c}}
    \toprule
    \multicolumn{2}{c|}{\textbf{Data Setting}} & \multicolumn{8}{c}{\textbf{Benchmark}} \\
    Multi-image & Single-image & Mantis-Eval & BLINK & MMIU & MathVista & MathVerse & MathVision & Remi & MV-MATH \\
    \midrule
    \xmark & \xmark & 70.80 & 55.23 & 52.00 & 68.20 & 46.30 & 25.07 & 36.38 & 20.41 \\
    \checkmark &\xmark & 74.54 & 57.84 & 53.70 & 68.5 & 46.1 & 26.60 & 39.12 & 20.58 \\    
    \xmark & \checkmark & 69.23 & 53.24 & 51.67 & 71.9 & 51.4 & 27.20 & 41.67 & 24.26 \\
    \checkmark & \checkmark & 76.39 & 58.53 & 54.23 & 73.0 & 49.56 & 28.26 & 41.65 & 25.68 \\
    \bottomrule
  \end{tabular}
  }
\end{table}
\paragraph{Single vs. Multi-Image}
Table~\ref{table:single_multi_combined_ablation} shows the impact of single-image and multi-image training data. Training exclusively on single-image datasets leads to limited generalization on multi-image benchmarks (e.g., significant drops on Mantis-Eval and BLINK), while training solely on multi-image data similarly reduces effectiveness on single-image reasoning benchmarks (e.g., MathVista). Interestingly, even without explicitly training on multi-image math reasoning data, performance improves on out-of-distribution benchmarks (Remi and MV-MATH), highlighting the model’s implicit learning of both inter-image relationships and reasoning capabilities. Moreover, MathVision benefits from multi-image training, likely due to its composite nature of multiple concatenated math reasoning images, which aligns closely with learned multi-image reasoning patterns.

% Dynamic sampling

%\paragraph{Policy update and Annealing}
% \paragraph{Different Architecture }
% We further evaluate the effectiveness of the proposed MMRL framework on different backbone architectures, namely Qwen2.5-VL-7B and InternVL2.5-8B. As illustrated in Table~\ref{table:architecture_ablation}, integrating MMRL consistently improves performance across both models and all benchmarks. Notably, Qwen2.5-VL-7B benefits more substantially from MMRL, exhibiting clearer gains such as a 7.41\% improvement on Mantis-Eval (from 68.98 to 76.39) and a 6.9\% increase on MathVista (from 68.2 to 73.0). These results indicate that while MMRL is generally effective across architectures, its benefits are more pronounced with Qwen2.5-VL, suggesting architecture-specific responsiveness to reinforcement-based augmentation strategies.

% \input{tables/different_architecture_ablation}

\section{Conclusion}
%In this paper, we aim to investigate multimodal reasoning of VLMs on general interleave multi-image tasks. We propose \OURS~model with interleaved multimodal reasoning through reinforcement learning. To deal with the challenges of the spatial and positional reasoning, we introduce a two-stage strategy to diversify the image order and resample the training data distribution for effective and efficient learning. Experimental results demonstrate the advanced performance on multi-image tasks and also the generalization to single-image tasks.

In this paper, we investigate the multimodal reasoning capabilities of VLMs on general interleaved multi-image tasks.  To address the challenges posed by spatial and positional reasoning, we introduce a multi-stage strategy that diversifies image order and resamples the training data distribution, enabling more effective and efficient learning. Experimental results demonstrate that our resulting model, Qwen-\OURS~not only achieves superior performance on multi-image benchmarks but also generalizes well to single-image tasks. As future work, we aim to investigate adaptive permutation strategies that can dynamically adjust image order based on task difficulty or model uncertainty.

%\section*{References}
\bibliographystyle{plain}
\bibliography{neurips_2025}

%%%%%%%%%%%%%%%%%%%%%%%%%%%%%%%%%%%%%%%%%%%%%%%%%%%%%%%%%%%%

\appendix

\newpage
\section{Technical Appendices and Supplementary Material}
\subsection{Training and Evaluation Details}
% TODO:fix on training details 

\begin{table}[!h]
\centering
\small  % 控制表格整体字号，与上面表格一致
\caption{Details of evaluation benchmarks.}
\label{tab:eval_bench_details}
\resizebox{1\columnwidth}{!}{
\begin{tabular}{l|c|c}
\toprule
\textbf{Benchmark} & \textbf{Description} & \textbf{\#samples} \\
\midrule
Mantis-eval   & Multi-image General Understanding QA & 217 \\
BLINK         & Multi-image General Understanding QA & 1901 \\
MMIU          & Multi-image General Understanding QA & 11698 \\
MathVista     & Single-image Math Reasoning QA       & 1000 (testmini) \\
MathVerse     & Single-image Math Reasoning QA       & 3940 \\
MathVision    & Single-image Math Reasoning QA       & 3040 \\
Remi          & Multi-image General Reasoning        & 2600 \\
MV-Math       & Multi-image Math Reasoning           & 2009 \\
\bottomrule
\end{tabular}}
\end{table}

We train our model on 8×H100 GPUs using the GRPO-based framework. The training batch size is 128, with a micro-batch size of 4 per GPU. The maximum prompt and response lengths are 8192 and 2048 tokens, respectively. We adopt Qwen2.5-VL-7B-Instruct\footnote{\url{https://huggingface.co/Qwen/Qwen2.5-VL-7B-Instruct}} as the backbone, enabling gradient checkpointing and FSDP without offloading. KL loss is applied with a coefficient of 0.01, while entropy regularization is disabled. The model is optimized for 2 epochs with a learning rate of $1 \times 10^{-6}$.

For evaluation, we use greedy decoding with temperature set to 0, top-p to 1, top-k to -1, and a maximum generation length of 2048. Evaluation benchmarks include MathVista, MathVerse, MathVision, and BLINK (configured via VLMEvalKit), while Mantis-Eval, MMIU, and MV-MATH (evaluated with official code via vLLM). All evaluations follow consistent decoding settings.Details are shown as Table~\ref{tab:eval_bench_details} and prompt ~\ref{prompt}

\subsection{Case study}
% TODO:Motivation reason(i.e different error reasons between qwen 7b and o4-mini-high on feishu)

\begin{figure}
% \vskip 0.2in
\begin{center}
\small
\centerline{\includegraphics[width=1.0\linewidth]{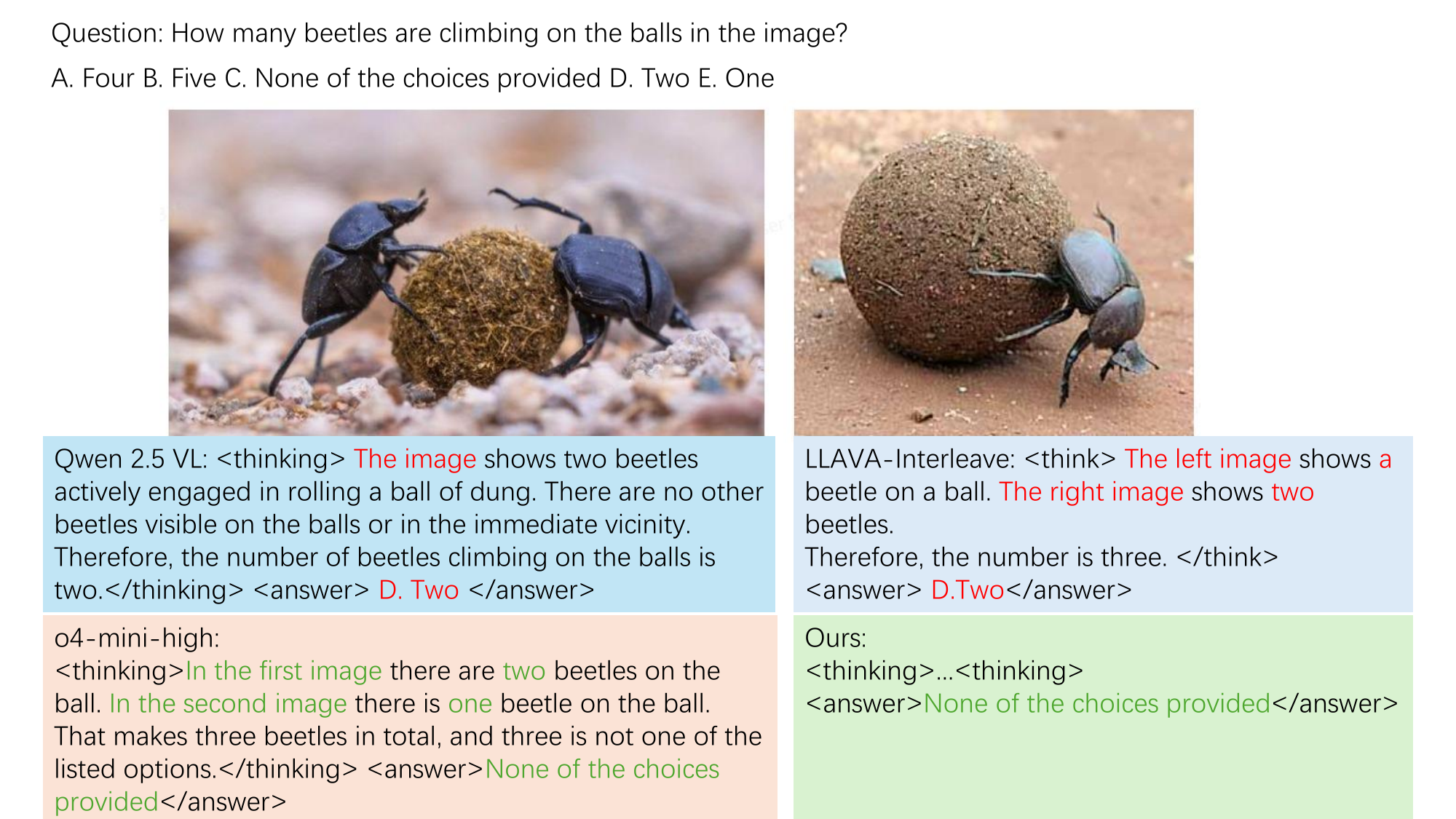}}
\caption{A showcase for motivation
   }
\label{fig:case_sup}
\end{center}
\vskip -0.2in
\end{figure}

As shown in Fig ~\ref{fig:case_sup}, we conduct a detailed comparison of the response quality across different types of models, including the proprietary model o4-mini-high, open-source base models, and previously trained interleaved models. We observe that the base models exhibit various types of errors, which can be broadly categorized into several patterns: missing relevant images, misinterpreting images, producing incorrect answers despite correct reasoning, and reaching incorrect conclusions despite accurate observations of details. For instance, in the beetle question shown above, Qwen frequently makes errors where the reasoning is correct but the final answer is wrong, or where some relevant visual information is overlooked. These are the types of failures we aim to address.

\begin{figure}[H]
\label{fig:case-study-1}
% 图片部分
\begin{minipage}[b]{0.3\linewidth}
  \includegraphics[width=\linewidth]{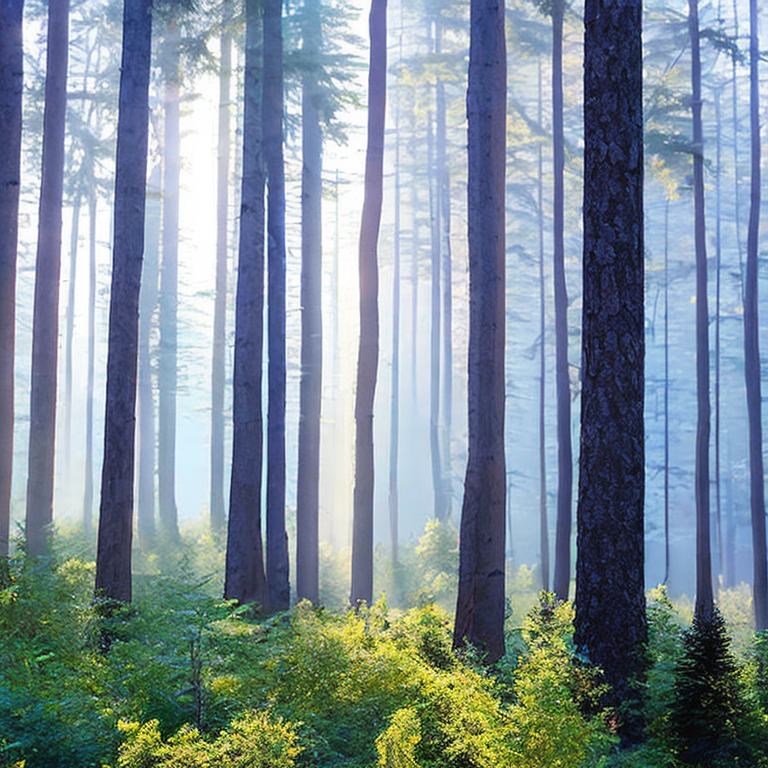}
  \centerline{\texttt{<image\_1>}}
\end{minipage}
\hfill
\begin{minipage}[b]{0.3\linewidth}
  \includegraphics[width=\linewidth]{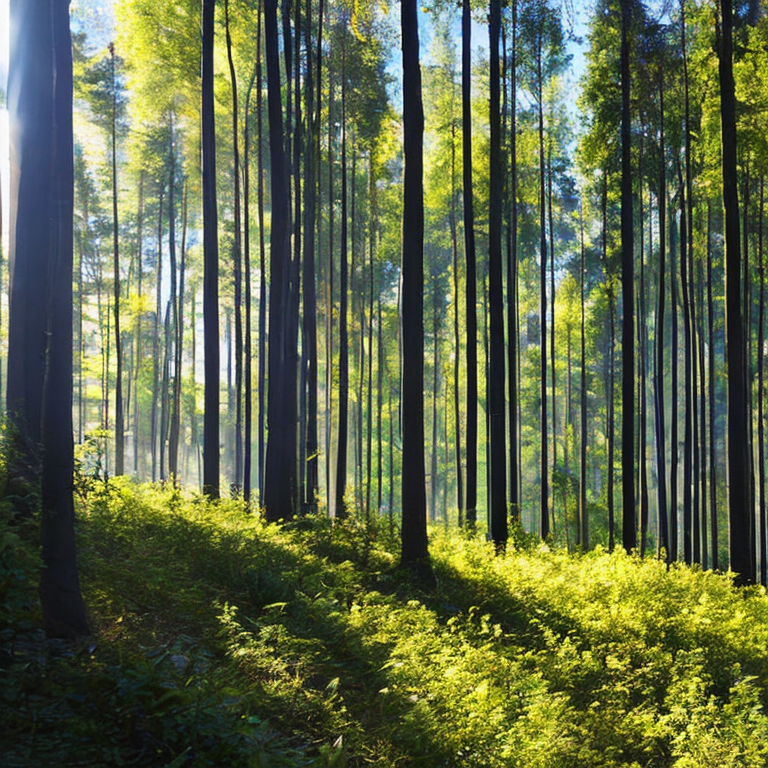}
  \centerline{\texttt{<image\_2>}}
\end{minipage}
\hfill
\begin{minipage}[b]{0.3\linewidth}
  \includegraphics[width=\linewidth]{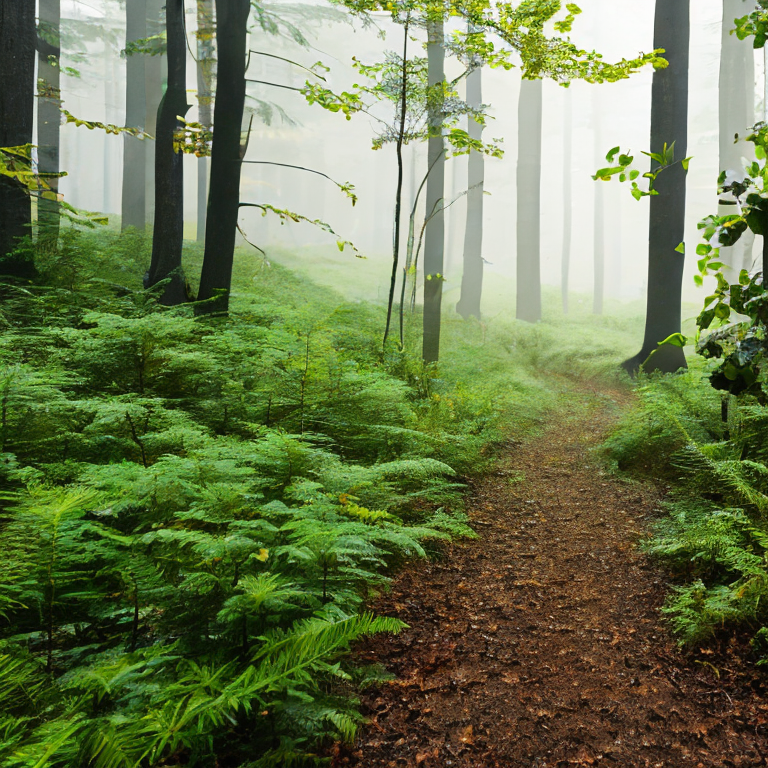}
  \centerline{\texttt{<image\_3>}}
\end{minipage}

% \vspace{-1em}

% % 第一个小问题框
\begin{tcolorbox}[
    colback=white!95!gray, 
    colframe=black, 
    title=Before Permuation,
    fonttitle=\bfseries\large
]
Which image is more similar to the reference image \texttt{<image\_1>}? Options: A. \texttt{<image\_2>} B. \texttt{<image\_3>}\\[0.5em]
\textbf{Qwen-2.5-VL-7B}: To determine which image is more similar to the reference image, let's analyze the key elements of both images:

\begin{itemize}[leftmargin=16pt, itemsep=2pt, parsep=0pt]
  \item Reference Image: 1) Features tall trees with a dense forest setting. 2) The lighting appears to be natural, possibly early morning or late afternoon, with sunlight filtering through the trees.

  \item Option A: 1) Also shows a forest scene with tall trees. 2) There is less mist or fog compared to the reference image.

  \item Option B: 1) Displays a forest path surrounded by lush greenery. 2) There is no mist or fog present, and the overall atmosphere is more open and less dense than the reference image.
\end{itemize}

Comparing these elements, Option A is closer to the reference image because it maintains the tall trees and dense forest setting. \checkmark

\end{tcolorbox}
% \end{figure}

% \begin{figure}[H]
% 第二个小问题框
\begin{tcolorbox}[
    colback=white!95!gray, 
    colframe=black, 
    title=After Permuation,
    fonttitle=\bfseries\large
]
Which image is more similar to the reference image \texttt{<image\_1>}? Options: A. \texttt{<image\_3>} B. \texttt{<image\_2>}\\[0.5em]
\textbf{Qwen-2.5-VL-7B}: To determine which image is more similar to the reference image, let's analyze the key elements of both images:

\begin{itemize}[leftmargin=16pt, itemsep=2pt, parsep=0pt]
  \item Reference Image: 1) Features tall trees with a dense forest setting. 2) The lighting suggests early morning or late afternoon, with sunlight filtering through the trees. 3) The ground is covered with green foliage, and there is a misty atmosphere.

  \item Option A: 1) Also features a forest with tall trees. 2) There is a path visible, which is not present in the reference image. 3) The lighting appears to be diffused, possibly due to the mist, but it is less pronounced than in the reference image. 4) The ground is covered with ferns and other greenery, similar to the reference image.

  \item Option B: 1) Features tall trees with a dense forest setting. 2) The lighting is bright and direct, suggesting midday sun. 3) The ground is covered with green foliage, but the overall scene lacks the misty atmosphere seen in the reference image.
\end{itemize}

Given this analysis, option B lacks the misty atmosphere and has brighter lighting, making it less similar to the reference image. The image that is most similar to the Reference Image is Option A. \xmark

\end{tcolorbox}

% \end{tcolorbox}
\caption{Case study demonstrating positional bias. The model correctly identifies the similar image when presented first (Before Permutation) but fails when the order is swapped (After Permutation).}
\label{fig:case1}
\end{figure}

% TODO:More analysis on failure cases on ours and base model
Besides, in case fig ~\ref{fig:case1}, we observe a consistent pattern in model behavior across the permutations of images: the model exhibits a stronger alignment between the reference image and the image presented first in the sequence of options. This suggests current VLMs have a positional bias in interleaved vision tasks.
\begin{figure}
% \vskip 0.2in
\begin{center}
\small
\centerline{\includegraphics[width=1.0\linewidth]{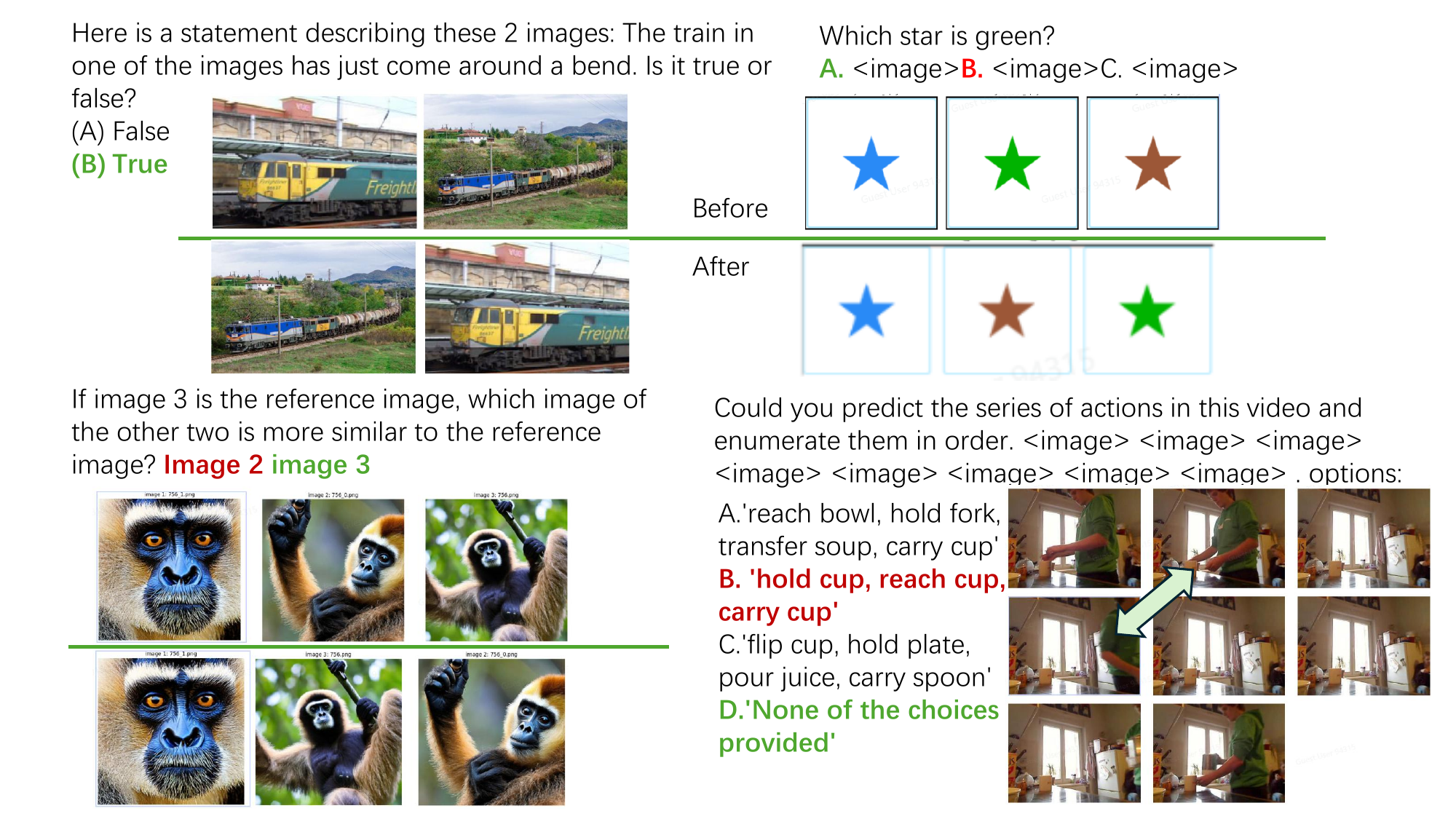}}
\caption{Training Data preprocessing pipeline explanation and several cases 
   }
\label{fig:case_data}
\end{center}
\vskip -0.2in
\end{figure}
% \begin{figure}[tb]
% % \vskip 0.2in
% \begin{center}
% \centerline{\includegraphics[width=1.0\linewidth]{figs/case/2/case-2.pdf}}
% \caption{A showcase for motivation
%    }
% \label{fig:case_sup}
% \end{center}
% \vskip -0.2in
% \end{figure}
% \begin{tcolorbox}[
%   boxrule=0pt, 
%   colframe=white, 
%   colback=white, 
%   sharp corners,
%   boxsep=0pt,
%   left=0pt,
%   right=0pt,
%   top=0pt,
%   bottom=0pt
% ]
% \label{box:case1}

\subsection{prompt} \label{prompt}
\subsubsection{Prompt for data preprocessing}

During the data preprocessing stage, we employed GPT-4o in conjunction with the prompts provided below to reformat questions and perform semantic variation checks.

\begin{tcolorbox}[
    colback=white!95!gray, 
    colframe=black, 
    title=Format Rephrase,
    fonttitle=\bfseries\large,
]

Your task is to convert a given QA conversation into a multiple-choice question format, and determine whether the question follows the given question type.

\vspace{6pt}

\textbf{\large Question Type Definition}

\vspace{3pt}

1) A question is considered a \textbf{Reference-Image Comparison} if it satisfies all of the following conditions:

\begin{itemize}[leftmargin=16pt, itemsep=2pt, parsep=0pt]
\item The question presents three or more images (e.g., "\texttt{<image> <image> <image>}").
\item One image is clearly identified as the \textbf{reference image} (e.g., "image 1 is the reference").
\item The question asks which of the remaining images is most similar to or most different from the reference image.
\item The answer options correspond only to the non-reference images.
\end{itemize}

2) \ldots

\vspace{6pt}

\textbf{\large Task Instructions}

\vspace{3pt}

\textbf{1) Convert the original QA pair into a multiple-choice question:}

\begin{itemize}[leftmargin=16pt, itemsep=2pt, parsep=0pt]
    \item Rephrase the assistant's response into an answer option (e.g., "A", "B", "C").
    \item Use placeholder tokens (\texttt{<image>}) in both the question and the options.
    \item Include only the images being compared (exclude the reference image from the options).
    \item Format the question strictly according to the example below.
\end{itemize}

\vspace{3pt}

\textbf{2) Determine the type of the question.}

\vspace{6pt}

\textbf{\large Output Format}

\vspace{3pt}

Return a JSON object with the following structure:

\vspace{3pt}

\begin{tcolorbox}[
    colback=gray!10!white,
    colframe=gray!50!black,
    boxrule=0.5pt,
    left=4pt,
    right=4pt,
    top=4pt,
    bottom=4pt,
    arc=2pt
]
\begin{verbatim}
{
  "question": "<multiple-choice question in specified format>",
  "answer": "<correct option letter>",
  "question_type": "<ReferenceComparison or Other>"
}
\end{verbatim}
\end{tcolorbox}

\vspace{6pt}

\textbf{\large Example}

\vspace{3pt}

\textbf{Input:} "Question: Answer the following question: Here are three images: \texttt{<image\_1>} \texttt{<image\_2>} \texttt{<image\_3>}. If image 1 is the reference image, which image of the other two is more similar to the reference image? Answer: The image that is more similar to the reference image is image 2."

\vspace{4pt}

\textbf{Output:}

\vspace{3pt}

\begin{tcolorbox}[
    colback=gray!10!white,
    colframe=gray!50!black,
    boxrule=0.5pt,
    left=4pt,
    right=4pt,
    top=4pt,
    bottom=4pt,
    arc=2pt
]
\begin{verbatim}
{
    "question": "Answer the following question: Which image 
        is more similar to the reference image <image_1>?
        Options: A. <image_2> B. <image_3>",
    "answer": "A",
    "question_type": "ReferenceComparison"
}
\end{verbatim}
\end{tcolorbox}

\end{tcolorbox}

\begin{tcolorbox}[
    colback=white!95!gray, 
    colframe=black, 
    title=Semantic Variation Check,
    fonttitle=\bfseries\large
]
You will be given a question involving one or more images, which are represented using image tokens. The token \texttt{<image>} represents an actual image. The tokens \texttt{<image\_1>}, \texttt{<image\_2>}, etc., refer to specific images by their positions (e.g., \texttt{<image\_1>} refers to the first image, \texttt{<image\_2>} to the second image).

\vspace{4pt}

\textbf{Your task is to determine the following:}

\vspace{2pt}

\begin{enumerate}[leftmargin=18pt, itemsep=3pt, parsep=0pt]
    \item If only the order of the images (e.g., \texttt{<image\_1>}, \texttt{<image\_2>}, etc.) is changed, would the answer to the question need to change?
    
    \item Is the question structured such that a single main image appears in the question body, and other images are referenced in the choices?
\end{enumerate}

\vspace{4pt}

\textbf{Respond in the following JSON format:}

\vspace{2pt}

\begin{tcolorbox}[
    colback=gray!10!white,
    colframe=gray!50!black,
    boxrule=0.5pt,
    left=4pt,
    right=4pt,
    top=4pt,
    bottom=4pt,
    arc=2pt
]
\begin{verbatim}
{
    "should_change": true or false,
    "is_multichoice_images": true or false
}
\end{verbatim}
\end{tcolorbox}

\vspace{2pt}

\textit{Note: This applies to both multiple-choice and fill-in-the-blank questions involving image references.}

\end{tcolorbox}

\subsubsection{Prompt for training and evaluation }
\label{eval-prompt}
\begin{tcolorbox}[
    colback=white!95!gray,
    colframe=black,
    title=Reasoning Format,
    fonttitle=\bfseries\large
]
\textbf{Instruction:} \\
You first think about the reasoning process as an internal monologue and then provide the final answer. \\
The reasoning process must be enclosed within \texttt{<think>} \texttt{</think>} tags. \\
The final answer must be put in \verb|\boxed{}|.
\end{tcolorbox}

% \paragraph{}
\subsection{Adavantage Differences}
% TODO:Difference on advantage during training stage
\begin{figure}
% \vskip 0.2in
\centering
\begin{subfigure}[t]{0.48\linewidth}
    \centering
    \includegraphics[width=\linewidth]{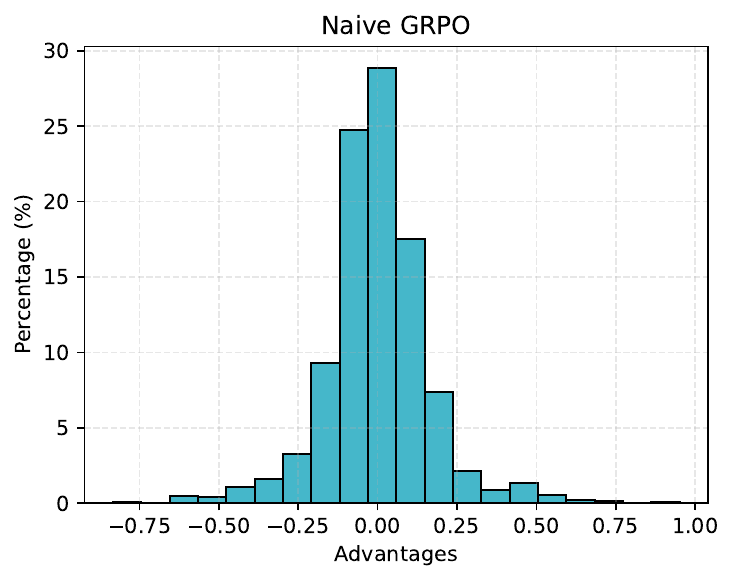}
    \caption{Naive GRPO}
    \label{fig:adv1}
\end{subfigure}
\hfill
\begin{subfigure}[t]{0.48\linewidth}
    \centering
    \includegraphics[width=\linewidth]{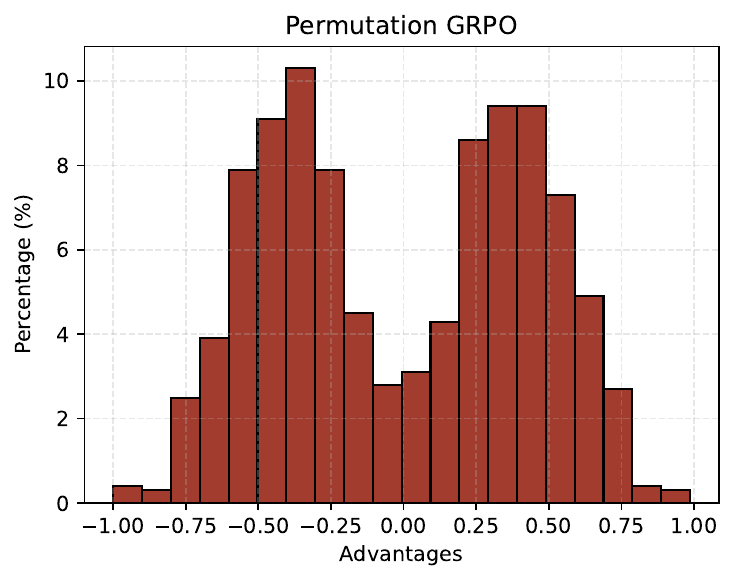}
    \caption{Permuation GRPO}
    \label{fig:adv2}
\end{subfigure}
\caption{Difference on advantage between Naive GRPO and Permuation GRPO during training stage.}
\label{fig:adv}
\vskip -0.2in
\end{figure}

As Fig~\ref{fig:adv} shows,the distinction becomes clearer when examining the advantage distributions during training. For multi-image inputs, Naive GRPO yields a distribution sharply peaked around zero, indicating that a large portion of training examples contribute negligible or ineffective gradient signals. In contrast, the permutation-based GRPO introduces greater input diversity by altering image order, which encourages the model to genuinely capture positional biases. This diversification leads to more informative advantage signals, facilitating more effective gradient updates during optimization.

% Technical appendices with additional results, figures, graphs and proofs may be submitted with the paper submission before the full submission deadline (see above), or as a separate PDF in the ZIP file below before the supplementary material deadline. There is no page limit for the technical appendices.

%%%%%%%%%%%%%%%%%%%%%%%%%%%%%%%%%%%%%%%%%%%%%%%%%%%%%%%%%%%%

\end{document}